\documentclass[11pt,oneside]{article}
\usepackage[utf8]{inputenc}
\usepackage[T1]{fontenc}
\usepackage[margin=1in]{geometry}
\usepackage{amsmath}
\usepackage[round,authoryear]{natbib}
\usepackage{url,microtype,subcaption,graphicx,xspace,booktabs,array,xcolor}
\usepackage[section]{placeins}
\usepackage[singlespacing]{setspace}
\usepackage[hidelinks]{hyperref}

\def\firstAuthorLast{Kharismawati and Kazic}

\def\Address{$^{1}$Department of Electrical Engineering and Computer Science, University of Missouri, Columbia, MO, USA \\
$^{2}$Missouri Maize Center, University of Missouri, Columbia, MO, USA \\
$^{3}$Interdisciplinary Plant Group, University of Missouri, Columbia, MO, USA \\
$^{4}$Department of Food, Agricultural and Biological Engineering, The Ohio State University, Columbus, OH, USA \\
$^{5}$MU Institute for Data Science and Informatics, University of Missouri, Columbia, MO, USA}
\def\corrAuthor{Dewi Endah Kharismawati}
\def\corrEmail{kharismawati.2@osu.edu}
\def\colmap{\textsc{COLMAP}\xspace}

\def\mesh{MeshLab\xspace}
\def\rgb{RGB\xspace}
\def\uav{UAV\xspace}
\def\uavs{UAVs\xspace}
\def\td{3D\xspace}
\def\exg{ExG\xspace}

\newcommand{\panel}[3]{\begin{minipage}[t]{#1}\centering#2\par\smallskip{\small #3}\end{minipage}}

\begin{document}
\title{Pointillist Maize: 3D Reconstruction of Field-Grown Maize Using \rgb \uav Imagery}
\author{Dewi Endah Kharismawati\textsuperscript{1,2,3,4,*}
\quad and \quad Toni Kazic\textsuperscript{1,2,3,5}\\[0.75em]
\small\textsuperscript{1}Department of Electrical Engineering and Computer Science,\\
\small University of Missouri, Columbia, MO, USA\\
\small\textsuperscript{2}Missouri Maize Center, University of Missouri, Columbia, MO, USA\\
\small\textsuperscript{3}Interdisciplinary Plant Group, University of Missouri, Columbia, MO, USA\\
\small\textsuperscript{4}Department of Food, Agricultural and Biological Engineering,\\
\small The Ohio State University, Columbus, OH, USA\\
\small\textsuperscript{5}MU Institute for Data Science and Informatics, University of Missouri, Columbia, MO, USA\\[0.5em]
\small\textsuperscript{*}Corresponding author: \href{mailto:kharismawati.2@osu.edu}{kharismawati.2@osu.edu}}
\date{}
\maketitle

\begin{abstract}
Three-dimensional reconstruction could extend field phenotyping from canopy summaries to the morphology of individual plants and organs, but field-grown maize is difficult to reconstruct because thin, flexible, mutually occluding structures move during image acquisition.
We investigated whether consumer-grade UAV RGB video processed using Structure from Motion and multi-view stereo could preserve recognizable maize structures, and how acquisition trajectories and frame sampling affected reconstruction characteristics.
COLMAP reconstructions preserved recognizable leaves, stems, tassel branches, ears, visible silks, and brace roots in individual field-grown plants.
To compare image sampling strategies, we reconstructed the same plant using two sets of 295 frames from the same UAV video: a single orbit sampled at 5 frames/second and three consecutive orbits at different altitudes (ribbon skirt) sampled at 2 frames/second.
After ground removal, the reconstructions contained 520,601 and 375,318 points, respectively.
The single orbit produced greater point concentration within the plant structure and fewer visible gaps, whereas the ribbon skirt produced fewer scattered points and slightly lower proportions of isolated points under the tested neighborhood thresholds.
In densely planted nursery rows, visible reconstruction voids occupied 17.57\% of the rendered scene using a single orbit at 1 frame/second, decreasing to 5.07\% with additional serpentine passes and 2.68\% when the original orbit was sampled at 3 frames/second.
However, additional observations also introduced greater point dispersion associated with plant movement.
A curlicue trajectory evaluated in a different field during the 2024 growing season produced distinguishable individual plants without visible duplicated structures, although substantial voids remained, primarily on the soil surface.
These findings demonstrate the feasibility of reconstructing recognizable maize organs from consumer UAV imagery, highlight trade-offs among acquisition strategies and reconstruction completeness, and provide a foundation for publicly accessible three-dimensional maize datasets, although independent geometric validation remains necessary for quantitative phenotyping.

\end{abstract}


\section{Introduction}
High-throughput phenotyping remains a major bottleneck in crop genetics and breeding, especially in field experiments where large populations must be observed repeatedly and at biologically relevant stages \citep{furbank2011,araus2014,minervini2015,hui2018}. 
Unoccupied aerial vehicles (UAVs) have expanded the spatial coverage, frequency, and flexibility of field data collection and enabled applications including multispectral crop-disease assessment \citep{liebisch2015,hu2018a,wang2019a,kharismawati2026c,liu2020a,zhu2020c,velumani2021,adak2023}. 
Much of this work summarizes plots or canopies through spectral response, vegetation indices, temperature, or height.  
These measurements are valuable for monitoring crop status, but canopy summaries do not directly represent the morphology and development of individual leaves, tassels, ears, brace roots, or damaged tissues. 
Those structures are often where the effects of genotype, development, disease, and environmental stress become visible.

Extracting reliable plant-level information from aerial imagery is itself a computer vision problem. 
Crop fields contain repeated rows and visually similar plants, while moving foliage, illumination changes, shadows, and soil backgrounds introduce additional appearance variability.  
In freely flown maize video, limited telemetry and the repetition of plants and rows also complicate frame registration and mosaic construction \citep{kharismawati2025a,aktar2020,kharismawati2020}.  
A successful field pipeline therefore depends not only on the downstream model, but also on what the acquisition makes observable, whether images can be registered consistently, and whether the representation preserves the spatial relationships required by the phenotype of interest.
This dependence on representation is also evident in maize stand counting,
where detections registered directly from raw video frames agreed more
closely with field counts than detections from mosaicked imagery \citep{kharismawati2025}.

Organ-level phenotyping particularly requires structural information.  
A two-dimensional image projects different depths onto the same plane: organs from adjacent plants may overlap, a front leaf may hide an ear or stem, and a visible fragment may be difficult to assign to its plant of origin.  
A three-dimensional representation can preserve relative position and orientation and can support organ segmentation and structural measurement \citep{li2022c,li2022d,yang2024}.  
This potential is especially relevant in maize.  
Its leaves are long, thin, curved, and flexible; tassels contain many fine branches; ears and silks may be partly concealed; brace roots occur near a cluttered soil boundary; and neighboring plants crowd each other as the canopy closes.  
The same morphology that makes a three-dimensional representation useful also makes it difficult to construct.
Recent studies demonstrate what can be achieved when acquisition is sufficiently controlled.  
Multi-view \rgb reconstruction has supported measurements of plant height, leaf length, leaf width, and relative leaf area in maize seedlings \citep{li2022c}.  
Organ-level point-cloud segmentation has enabled extraction of leaf inclination, leaf growth height, and related structural traits from individual maize shoots \citep{li2022d}.   
Larger annotated maize point-cloud datasets further demonstrate the importance of preserving stem and leaf geometry for automated analysis across diverse architectures \citep{yang2024}.   
LiDAR, controlled multi-camera systems, and turntable imaging can likewise provide detailed plant models \citep{luo2021a,yin2020,lin2021,gollob2021,zhu2020a}.   
These results establish the value of plant-scale \td data, but many of these approaches rely on stationary or potted plants, structured imaging platforms, controlled backgrounds, or carefully constrained camera motion. 
Such conditions reduce the challenges associated with plant movement, occlusion, and variable illumination, but are difficult to reproduce at the scale of an open agricultural field. 
Consequently, methods that perform well under controlled conditions may not scale readily to field environments, where large plant populations and dense canopies make it difficult to capture the structural details of individual plants.
Field-grown plants present several challenges for conventional three-dimensional reconstruction methods.   
Structure from Motion (SfM) estimates camera poses from correspondences among images, while multi-view stereo densifies the recovered scene \citep{schoenberger2016,schoenberger2016a}.   
Both are most straightforward when the scene is rigid, distinctive visual features persist across views, and each surface is observed repeatedly \citep{schoenberger2016,schoenberger2016a}.   
Maize has slender organs and extensive self-occlusion, which can produce incomplete leaves and missing tips in multi-view reconstructions \citep{wu2022a,li2022c}.   
Wind can move individual organs while images are recorded, and variable illumination can change their appearance between views \citep{wu2022a,li2022c}. %
At range scale, overlapping leaves and increasingly dense canopies increase occlusion and make individual plants more difficult to separate \citep{zermas2020,sweet2022}.   
Consequently, a failure may reflect missing observations, nonrigid movement, ambiguous correspondence, insufficient image resolution, or a combination of these factors rather than simply an inadequate number of frames.

UAV reconstruction has been successful for field structure and plant height estimation, but it has often emphasized canopy-scale surfaces rather than individual organs \citep{zermas2020}.   
Acquisition geometry is critical because a reconstruction cannot recover a surface that was never observed.   
In field maize canopies, oblique \uav photography produced more complete vertical profiles than nadir photography and recovered substantially more points below the upper canopy \citep{che2020}.   
A crop-surface-model study similarly found that oblique views improved maize reconstruction relative to nadir views and that reduced spatial resolution lowered plant height accuracy \citep{niu2024a}.   
Controlled multi-view platforms impose overlap requirements within and between camera levels and report incomplete leaves and tips when coverage is insufficient; wind and variable illumination remain important limitations when such systems move outside controlled environments \citep{wu2022a}.   
Together, this literature indicates that camera placement, overlap, distance, and view angle determine which structural evidence is available for reconstruction.

This creates a field acquisition trade-off that is not captured by frame count alone.   
Sampling many adjacent frames from a video can increase temporal redundancy and local point density, but closely spaced frames may contribute nearly the same evidence and cannot expose a persistently occluded surface \citep{gibbs2020}.   
Changing elevation or camera angle can reveal surfaces that are hidden in nadir or single-level views \citep{che2020,niu2024a,wu2022a}.   
Longer acquisitions also increase exposure to wind-driven motion of flexible plant organs, which can reduce consistency among views \citep{wu2022a,kharismawati2026}.   
Flying farther away increases coverage but decreases spatial resolution and the number of pixels representing a leaf edge, tassel branch, or silk \citep{niu2024a}.   
Thus, temporal sampling, viewpoint diversity, target resolution, acquisition duration, and scene motion interact.   
A fair comparison must hold at least the source sequence and image budget constant rather than equating more frames or more reconstructed points with better evidence.

Given the prior work, it is not obvious that plants in field environments can be reconstructed sufficiently for organ phenotyping.
We therefore undertook a qualitative proof-of-concept study to determine the feasibility of plant reconstructions for different planting situations in the field, moving from isolated plants to the standard nursery planting of rows and ranges.
Here, we show that plants in field environments can be reconstructed in three dimensions for organ phenotyping from video shot from consumer-grade \uavs.  We explore different planting schemes and trajectories and examine the trade-offs between these, reconstruction quality and resolution, and acquisition times.
Established software such as \colmap makes SfM and multi-view stereo accessible without requiring a new reconstruction algorithm \citep{colmap,schoenberger2016}.   
The unresolved question is what a 
conventional and reproducible workflow can recover from \uav video of mature maize in the field and what limits emerge as acquisition scales up from one to many plants.   
This is important for researchers designing data acquisition strategies, whether with their existing \uavs or any prospective purchase of specialized, more complex instruments.
%
%
%

%
%
%
This study makes three contributions.
First, it demonstrates the feasibility of imaging unstructured fields and the quality of the three-dimensional reconstructions of maize plants and their organs, including leaves, tassels, ears, silks, and brace roots.
%
%
%
Second, it compares two image sampling strategies, 
%
%
examining
how different combinations of temporal sampling and acquisition
viewpoints affect 
%
%
how the points are distributed in the three-dimensional point cloud relative to the plants' structures.
Third, it investigates how the plants' structural details change with flight trajectories that capture larger and more complex scenes.
%
%
%
Together, these findings provide practical guidance for
acquiring and processing \uav imagery for \td reconstruction
and support the longer term goal of developing publicly
accessible \td datasets of maize grown under field conditions.
%
%
%
%
%
%

%
%
%
\section{Materials and Methods}

\subsection{Field Site and Plant Material}
Maize (\emph{Zea mays}) was grown in Columbia, Missouri, USA, during the 2022 and 2024 field season.   
The material included three inbreds, an elite line, and approximately 100 near-isogenic disease lesion mimic mutant lines.  
The field contained two relevant layouts.   
In a tiling section, individual plants were placed at vertices of nearly equilateral triangles with sides of approximately 4.5 m.   
Five plants each of an inbred, an elite line, and one lesion mimic mutant (\emph{Mo20W/les$^\ast$-N1395C}) were planted in this section.   
The separation permitted imaging around an individual plant with limited interference from neighboring plants.   
In the genetic nursery, rows were 0.9 m apart and plants were spaced 24.5 cm apart within rows.   
A range is a set of nursery rows bounded by unplanted alleys.   
The same field layout was maintained during the 2024 growing season, with the addition of 14-inch pots installed around individual plants to improve watering efficiency.

\subsection{UAV Imaging and Flight Configurations}
Video was acquired manually with a 2021 DJI Mavic 2 Pro carrying its \rgb camera.   
The source videos were recorded at 30 frames/second and $3840\times2160$ pixels. 
%
Flights were designed to capture diverse viewpoints of plant surfaces rather than follow predefined automated flight paths.

\subsubsection{Individual Plant Acquisition}
Figure~\ref{fig:equalframes} shows two different orbital trajectories for acquiring individual plants.  In the top panel of  Figure~\ref{fig:equalframes}(a) we show a single-orbit trajectory, in which the \uav completed one full circuit around the plant at one flight level, at an approximate radius of 1--2 m, and different camera angles. 
The bottom panel of Figure~\ref{fig:equalframes}(a) shows a multi-level trajectory that we define here as a \emph{ribbon skirt}: three completed, vertically stacked orbits around the same plant, with the flight altitude and camera angle changed between orbits.
The \uav completed each full orbit at a fixed level before increasing its altitude and beginning the next orbit.  
During the first ribbon skirt orbit, the \uav flew at approximately half the plant height with a parallel camera pose, defined here as a dead-on view with the camera optical axis approximately parallel to the ground.  
During the second orbit, the \uav flew at approximately the height of the plant and the camera was angled slightly downward to provide an oblique view.  
During the third orbit, the \uav flew approximately 2 m above the top of the plant and the camera was angled more steeply downward to provide a sharply oblique view.  
Thus, the \uav ascended only between completed orbits; the ribbon skirt was not a continuously ascending spiral.  
The \uav was flown manually, and the 1--2 m orbit radius, flight altitudes, and viewing angles are approximate rather than surveyed values.

\subsubsection{Range Level Acquisition}
Ranges were imaged at greater distance using three distinct trajectory types: an orbit around the range with an oblique camera pose, serpentine passes approximately perpendicular to the rows with oblique or nadir views, and a curlicue path.  
Curlicue denotes a continuous looping pattern with a smaller radius at approximately constant altitude.  
It was used for a range containing multiple rows and is distinct from the multi-level individual plant acquisition.  
Source descriptions give approximate altitude ranges of 5--20 m for range orbits and 3--10 m for serpentine passes.  
The principal range example was acquired at approximately 5--7 m.  
These values describe the broader collection rather than a controlled single-factor comparison.
The 2024 curlicue experiment was conducted in a different field during a subsequent growing season, while maintaining the same planting layout as the 2022 nursery range experiments.

\subsection{Video Frame Extraction and Experimental Configurations}
Frames were extracted from video at the sampling rates listed in Table~\ref{tab:configurations}. Comparisons of different flights are descriptive because plant material, flight altitude, camera geometry, scene complexity, and environmental motion can vary together.

The controlled equal-frame comparison used one source video of the same individual maize plant.  
Configuration A comprised 295 frames from one orbital level and viewing configuration, sampled at 5 frames/second.  
Configuration B comprised 295 frames spanning three consecutive complete orbits at different elevations and angles, sampled at 2 frames/second.  
Thus total frame count and source video were held constant, while the distribution of viewpoints differed.  
Frame sampling rate and viewpoint diversity are therefore separate experimental factors.

\begin{table}[!htbp]
\centering\scriptsize
\setlength{\tabcolsep}{3.2pt}
\begin{tabular}{llp{3cm}rrl}
\toprule
\textbf{Scene} & \textbf{Trajectory} & \textbf{camera pose} & \textbf{frames/second} & \textbf{frames} & \textbf{purpose}\\
\midrule
plant & one orbital level & single level/view & 5 & 295 & equal-frame comparison\\
plant & three complete orbits & parallel to oblique & 2 & 295 & equal-frame comparison\\
plant & two orbits & parallel & 1 & 177 & wild-type example\\
plant & two orbits & parallel & 1 & 129 & mutant example\\
plant & two orbits & parallel and oblique & 2 & 219 & pose and masking examples\\
tiling & one orbit & oblique & 1 & 150 & separated plant group\\
range & one orbit & oblique & 1 & 126 & range example\\
range & orbit plus serpentine & oblique and nadir & 1 & 208 & trajectory comparison\\
range & one orbit & oblique & 3 & 426 & sampling comparison\\
range (2024) & curlicue & oblique & $\sim$0.5 & 287 & curlicue reconstruction \\

\bottomrule
\end{tabular}
\caption{Verified video subsets represented in the manuscript sources. The two 295-frame configurations came from the same source video. Other rows are not controlled single-factor comparisons. Frames/second is the rate at which frames were extracted from the 30 frames/second source video.}
\label{tab:configurations}
\end{table}

\subsection{\td Reconstruction and Postprocessing}
Extracted frames were processed with \colmap using one NVIDIA RTX 2080 Ti GPU in a computer with two RTX 2080 Ti GPUs and 128 GB RAM. \colmap detected and matched image features, estimated camera poses with SfM, calculated depth with multi-view stereo, and fused the depth estimates into dense point clouds, which were exported as PLY files \citep{colmap,schoenberger2016}.

Some image sets were masked before reconstruction. One approach transformed \rgb frames to the device-independent \emph{L*a*b*} color space, where L* represents lightness, a* represents the green to red color axis, and b* represents the blue to yellow color axis. Pixels were selected based on their values in these color channels. A combined mask retained pixels satisfying
\begin{equation}
\theta=[(a^*<0)\wedge(b^*>0)]\vee[(a^*>0)\wedge(b^*<0)],
\label{eq:lab}
\end{equation}
where $\theta$ represents the resulting mask, $<$ and $>$ indicate values
below and above the specified thresholds, respectively, $\wedge$ means
that both conditions must be satisfied, and $\vee$ means that either
group of conditions may be satisfied.
The first group selects pixels with green and yellow color components,
while the second selects pixels with red and blue color components.
The latter was intended to retain some tassel regions but could also
include sky pixels. 
A more restrictive experimental mask used
\begin{equation}
\theta=(L^*<35)\wedge(a^*<-20)\wedge(b^*>10).
\label{eq:hardmask}
\end{equation}
This mask retained only pixels that simultaneously had a lightness
value below 35, an $a^*$ value below $-20$ (toward green), and a
$b^*$ value above 10 (toward yellow).
These threshold values were determined empirically through
repeated visual inspection and adjustment of the resulting masks.
The selected values provided the best observed retention of plant
structures, including leaves, tassels, and ears, among the
threshold combinations tested.
RGB frames were also masked using excess green,
\begin{equation}
\mathrm{ExG}=2G - R - B,
\label{eq:exg}
\end{equation}
where $R$, $G$, and $B$ represent the red, green, and blue color
channel values, respectively.
Excess green emphasizes vegetation by increasing the contribution
of green relative to red and blue.
The resulting values were normalized to the range $[0,1]$, followed
by Otsu thresholding to automatically separate pixels into two
groups based on their excess green values.
Thresholds were selected by inspecting whether masks retained leaves, tassels, and ears while excluding soil and sky. 
These methods were exploratory and were not trained segmentation models.

Point clouds were inspected and postprocessed in \mesh \citep{cignoni2008}.  
Cropping denotes restricting a reconstruction to a region of interest.  
Manual ground cleaning denotes selection and removal of soil and ground points.  
For the equal-frame comparison, approximately the same manual ground-removal procedure was applied to both clouds.  
Above-ground points around leaves, tassels, and other structures were not manually removed.  
The reported 520,601 and 375,318 points are therefore counts after soil and ground removal, not counts from the original unprocessed \colmap clouds and not accuracy metrics.

Surface reconstruction was explored with Poisson and ball pivoting methods.  
The most useful visual results in these experiments came from MeshLab's automatic ball-radius estimate with a clustering radius equal to 10\% of the ball radius; one manually edited example used 5\% \citep{bernardini99}.  
These meshes were visualizations of the point clouds, not geometric validation. 
Manual above-ground cleaning was performed only for the separate surface-reconstruction demonstration and must not be confused with the equal-frame comparison.
\subsection{Evaluation}
Reconstructions were evaluated through visual inspection of recognizable morphological structures, apparent structural continuity, scattered points, missing surfaces, duplicated structures, and misregistration of soil or plants.
Point counts were used to describe retained samples after specified processing steps, while local point density and isolation were quantified for the two configurations compared using an equal number of frames.
The evaluation focused on visual and spatial characteristics of the reconstructions because independent physical \td references and validated organ measurements were not available.
For the equal-frame comparison, local point density and isolation were also quantified directly from the two post-ground-removal PLY files using Python 3.9, NumPy 2.0.1, and SciPy 1.13.1. 
To define a common spatial scale, $s$, we calculated the pooled median Euclidean nearest-neighbor distance across both point clouds using a k-d tree ($s = 0.001279$ coordinate units).
Then for each point $i$, the number of neighboring points within a
fixed radius $r$ was calculated as
\begin{equation}
%
N_i(r) = \big| \left\{j\ne i: \lVert \mathbf{x}_j - \mathbf{x}_i \rVert_{\scriptscriptstyle 2} \le r \right\} \big|,
%
%
\label{eq:neighbors}
\end{equation}
where the outer vertical bars denote the number of points
satisfying the distance condition, $\mathbf{x}_i$ and
$\mathbf{x}_j$ represent the three-dimensional coordinates
of points $i$ and $j$, respectively, and
$r\in\{2s,3s,5s,10s\}$.
The median number of neighboring points and the proportions of
points having zero, at most two, or at most five neighbors were
calculated to characterize point concentration and isolation.
To determine whether the results depended on differences in
sampling density, the nearest neighbor distance for each point
was also divided by the median nearest neighbor distance within
its respective cloud.
Similarly, the mean distance to the eight nearest neighboring
points was divided by the corresponding median within each cloud,
and the proportions of points exceeding predefined multiples
of these normalized distances were reported.
These measurements characterize the spatial distribution of
points, distinguishing densely concentrated points within the
reconstructed plant structure from sparse or isolated points
in peripheral regions of the reconstructions.
However, one or more isolated points are not necessarily a reconstruction
artifact, as they may represent a legitimate leaf edge, tip,
tassel branch, or an incompletely observed surface, or an organ moving during acquisition.
No statistical significance test was performed at the individual
point level because the two point clouds represent single
experimental units, and their constituent points are not
independent biological or acquisition replicates.

For nursery range reconstructions, we quantified visible reconstruction voids using snapshots of the rendered point clouds shown in Figure~\ref{fig:range}.
Void regions were identified by their yellow color in the rendered images.
For the three reconstructions from the 2022 field experiment, snapshots were selected from the same nursery location and viewing region to ensure a consistent comparison.
For the curlicue reconstruction acquired in a different field during the 2024 growing season, a snapshot of a comparable nursery region was selected.
The visible void area percentage was calculated as the number of yellow pixels corresponding to reconstruction voids divided by the total number of pixels in each snapshot, multiplied by 100.
This measurement represents the projected extent of visible reconstruction gaps rather than the three-dimensional area or volume of missing geometry.
Because the curlicue experiment was conducted in a different field and growing season, its void percentage was interpreted descriptively rather than as a direct performance comparison with the 2022 reconstructions.

%


\section{Results}

\subsection{Feasibility of Individual Plant Reconstruction}
\label{sec:individual}
\begin{figure}
\centering
\panel{0.25\textwidth}
      {\includegraphics[width=\linewidth]{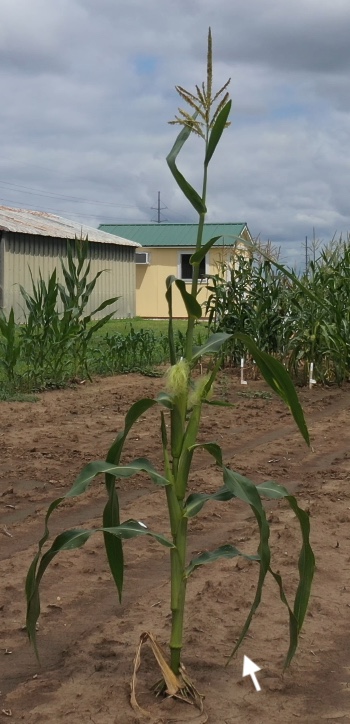}}
      {(a) Representative \rgb frame.}
\panel{0.35\textwidth}
      {\includegraphics[width=\linewidth]{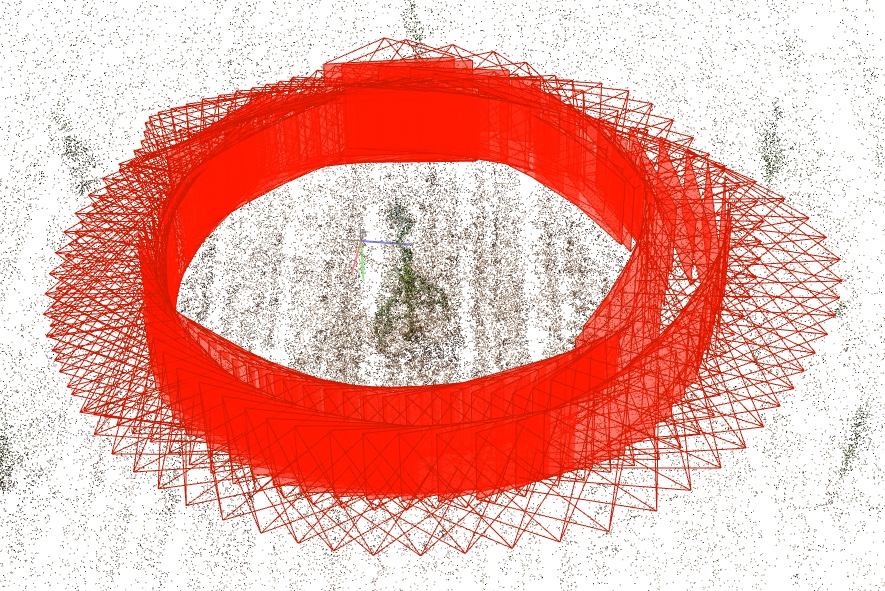}}
      {(b) Estimated camera poses and sparse point cloud.}
\panel{0.25\textwidth}
      {\includegraphics[width=\linewidth]{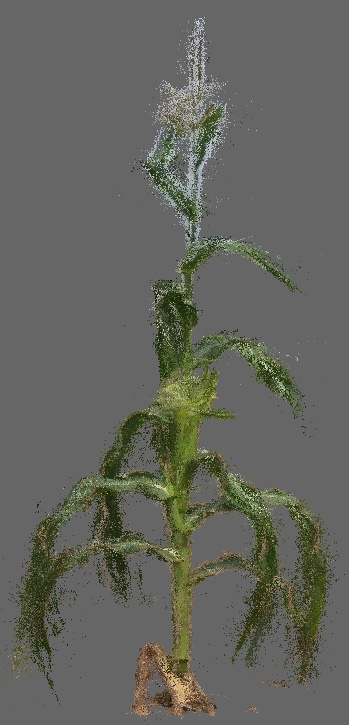}}
      {(c) Dense point cloud.}

\caption{Three-dimensional reconstruction of an
individual W23 maize plant using two orbital
trajectories with the same parallel camera orientation.
(a) Representative \rgb frame showing the
field-grown plant.
(b) Estimated camera poses and sparse point cloud
from the two orbital acquisitions.
(c) Dense point cloud showing recognizable
plant architecture, including the stem, curved
leaves, brace roots, a senescent lower leaf,
and the tassel.}

\label{fig:individual_parallel}
\end{figure}

The first question was whether consumer \uav \rgb video
could preserve recognizable maize morphology in an
isolated field-grown plant.
For the W23 plant, two orbital trajectories were
acquired with the camera oriented approximately
parallel to the ground (Figure~\ref{fig:individual_parallel}a,b).
The estimated camera poses surrounded the plant,
providing overlapping observations from multiple
directions.
The resulting dense point cloud preserved the overall
plant architecture, including the stem, curved leaves
with different orientations, brace roots, a senescent
lower leaf, and the tassel
(Figure~\ref{fig:individual_parallel}c).
Individual leaves remained visually distinguishable,
including those that crossed or partially occluded
one another in the original images.
The initial dense reconstruction contained 17,009,512
points, of which 1,990,141 remained after cropping
to the plant region and 613,244 remained after
removing most of the surrounding soil.
The reconstruction also contained scattered points,
particularly around leaf tips and the tassel,
where movement during acquisition affected the
consistency of observations.
Points near the plant boundaries sometimes retained
colors associated with the surrounding soil or sky.
Despite these artifacts, the reconstructed point
cloud preserved recognizable vegetative and
reproductive structures of the W23 plant.

\begin{figure}[t!]
\centering
\panel{0.16\textwidth}
      {\includegraphics[width=\linewidth]{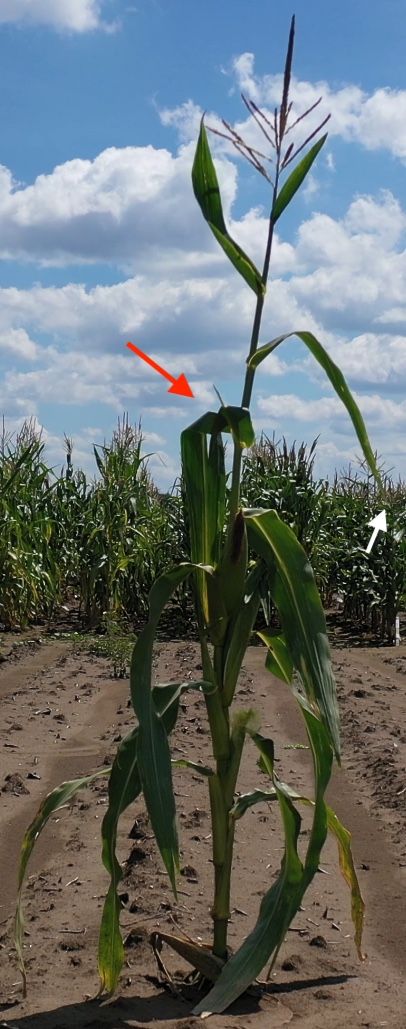}}
      {(a) Representative \rgb frame.}
\panel{0.27\textwidth}
      {\includegraphics[width=\linewidth]{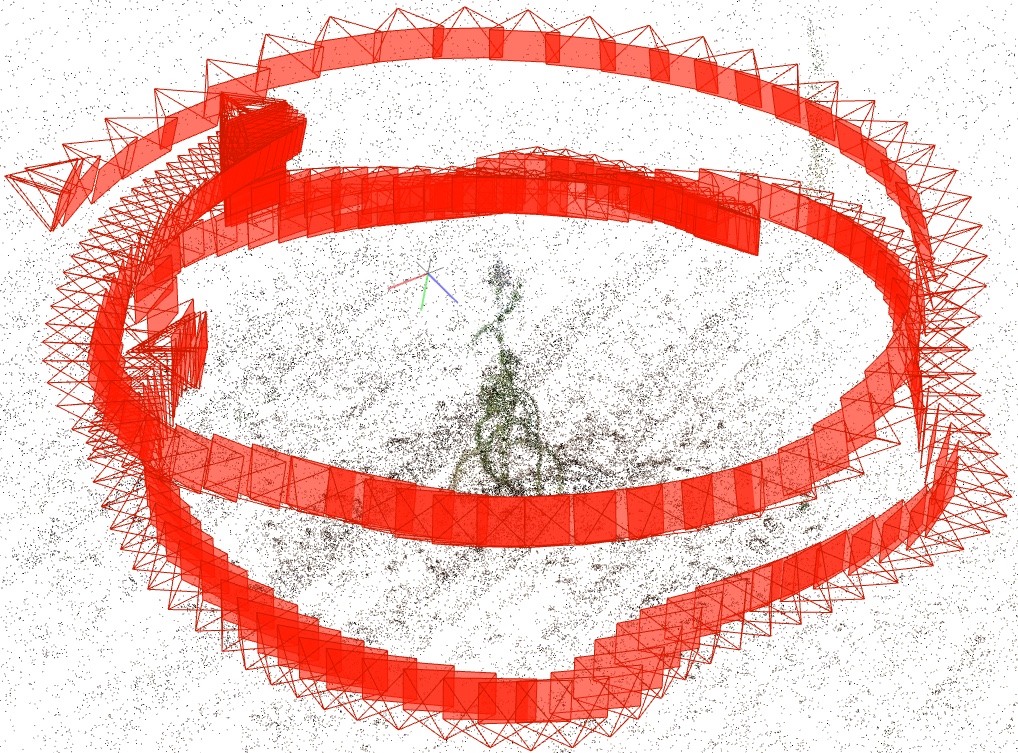}}
      {(b) Estimated camera poses and sparse point cloud.}
\panel{0.19\textwidth}
      {\includegraphics[width=\linewidth]{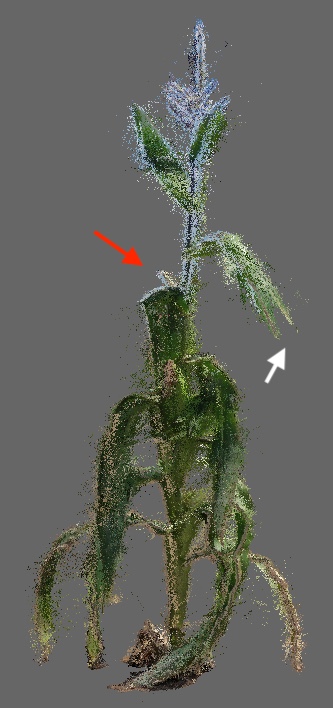}}
      {(c) Dense point cloud.}
      
\panel{0.20\textwidth}
      {\includegraphics[width=\linewidth]{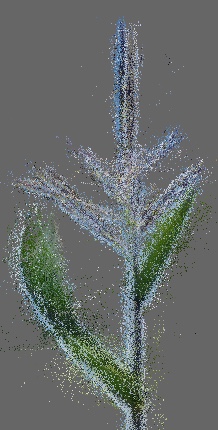}}
      {(d) Tassel detail.}
\panel{0.232\textwidth}
      {\includegraphics[width=\linewidth]{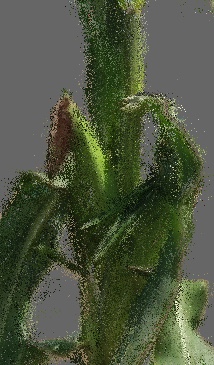}}
      {(e) Ear and silk detail.}
\panel{0.263\textwidth}
      {\includegraphics[width=\linewidth]{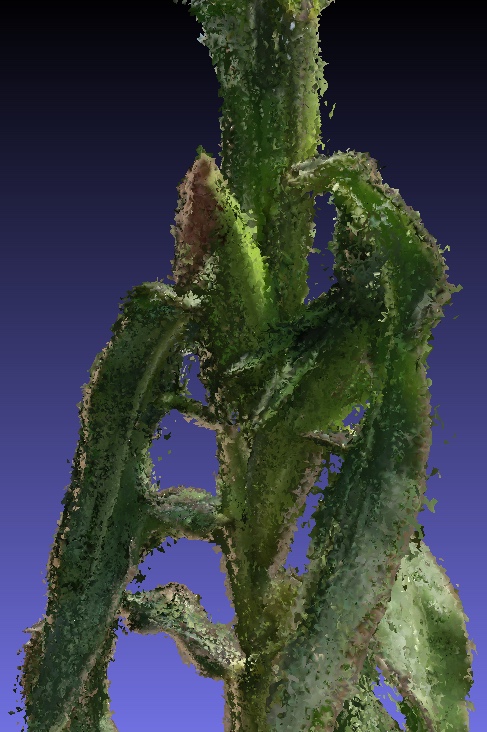}}
      {(f) Ball pivoting surface at the ear region.}
\panel{0.256\textwidth}
      {\includegraphics[width=\linewidth]{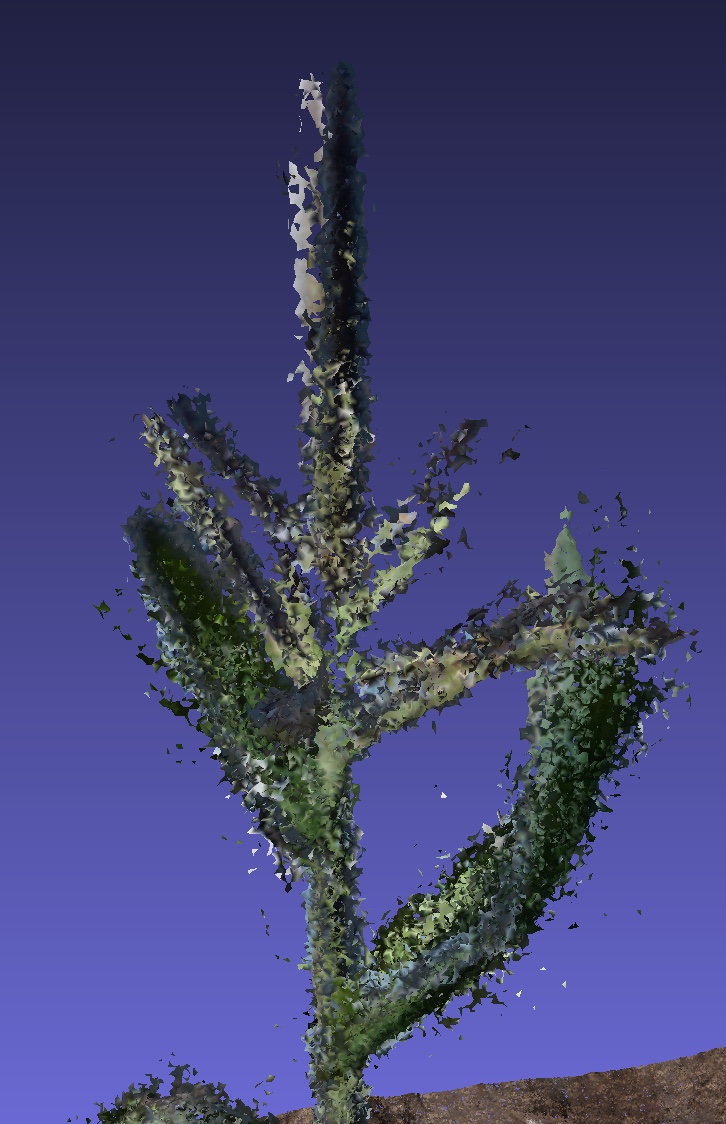}}
      {(g) Ball pivoting surface at the tassel region.}

\caption{Three-dimensional reconstruction of an
individual maize plant using two orbital
trajectories with parallel and oblique
camera orientations.
(a) Representative \rgb frame. 
%
(b) Estimated camera poses and sparse point cloud
showing the two orbital acquisitions.
(c) Dense point cloud showing the reconstructed
plant morphology.
%
(d,e) Detailed views of the reconstructed
tassel, ear, and silks.
(f,g) Surface meshes generated using the
ball pivoting algorithm, showing recognizable
structures in the ear and tassel regions
alongside areas of incomplete surface continuity.
In both (a) and (c), the red arrows identify a leaf torn along the midrib, with the broken midrib protruding above the leaf blade, while the white arrows identify a leaf that moved considerably during acquisition.}

\label{fig:individual_parallel_oblique}
\end{figure}

A second plant was reconstructed using two orbital
trajectories with different camera orientations.
The first orbit used a camera orientation approximately
parallel to the ground, while the second used an
oblique orientation to capture additional viewpoints
of the upper plant
(Figure~\ref{fig:individual_parallel_oblique}a,b).
The parallel orbit was acquired at approximately
0.6~m above the ground, while the oblique orbit
was acquired at approximately 3~m with a camera
angle of 20--30 degrees.
COLMAP processed 219 video frames and generated
25,010,235 points, with 1,288,682 points remaining
after removing the surrounding background and soil.
The resulting dense point cloud preserved the
overall plant architecture and finer reproductive
structures, including individual tassel branches,
an ear, and visible silks
(Figure~\ref{fig:individual_parallel_oblique}c--e).
The oblique viewpoints provided additional observations
of the upper plant and tassel, which were less
accessible from the lower parallel camera orientation.

Plant movement during orbital acquisition introduced
additional structures that did not correspond to
distinct physical organs.
Leaves changed position between observations in
response to ambient wind and air currents generated
by the \uav.
These changes violated the static-scene assumption
underlying the reconstruction and produced duplicated
or displaced leaf structures, referred to here as
ghost organs.
One example is visible in
Figure~\ref{fig:individual_parallel_oblique}c
(white arrow), where a moving leaf was reconstructed
in two different positions.
Movement was particularly apparent in younger upper
leaves and leaf tips, which are more flexible than
the larger and more rigid plant structures.
The resulting ghost organs introduced ambiguity
into the reconstructed leaf geometry even though
the overall plant architecture remained recognizable.

To obtain a surface representation of the reconstructed
plant, the ball pivoting algorithm in MeshLab was
applied to the dense point cloud.
The algorithm connected neighboring points with
triangular faces, converting the point-based
representation into a surface mesh.
The resulting surfaces preserved recognizable
morphological features, including leaf midribs,
ear husk leaves, visible silks, and tassel branches
(Figure~\ref{fig:individual_parallel_oblique}f,g).
Some fine structures, particularly the silks,
became easier to distinguish in the surface
representation than in the original point cloud.
However, regions with incomplete or inconsistent
point coverage contained holes and irregular
surface connections, particularly around thin
and complex organs.
The ball pivoting reconstruction therefore provided
a surface representation of the maize plant, but
did not recover missing geometry or eliminate
motion-related artifacts already present in the
underlying point cloud.

\subsection{Equal Frame Budget, Different Viewpoint Distributions}
\label{sec:equalframes}

\begin{figure}[t!]
\centering
\panel{0.43\textwidth}
{\includegraphics[width=\linewidth]{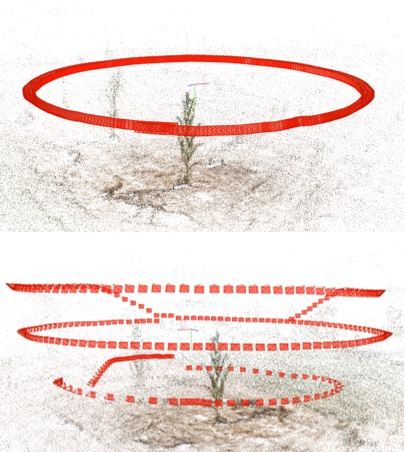}}
{(a) Flight trajectories: top, single orbit at 5 fps;
bottom, ribbon skirt at 2 fps.}
\hfill
\panel{0.265\textwidth}
{\includegraphics[width=\linewidth]{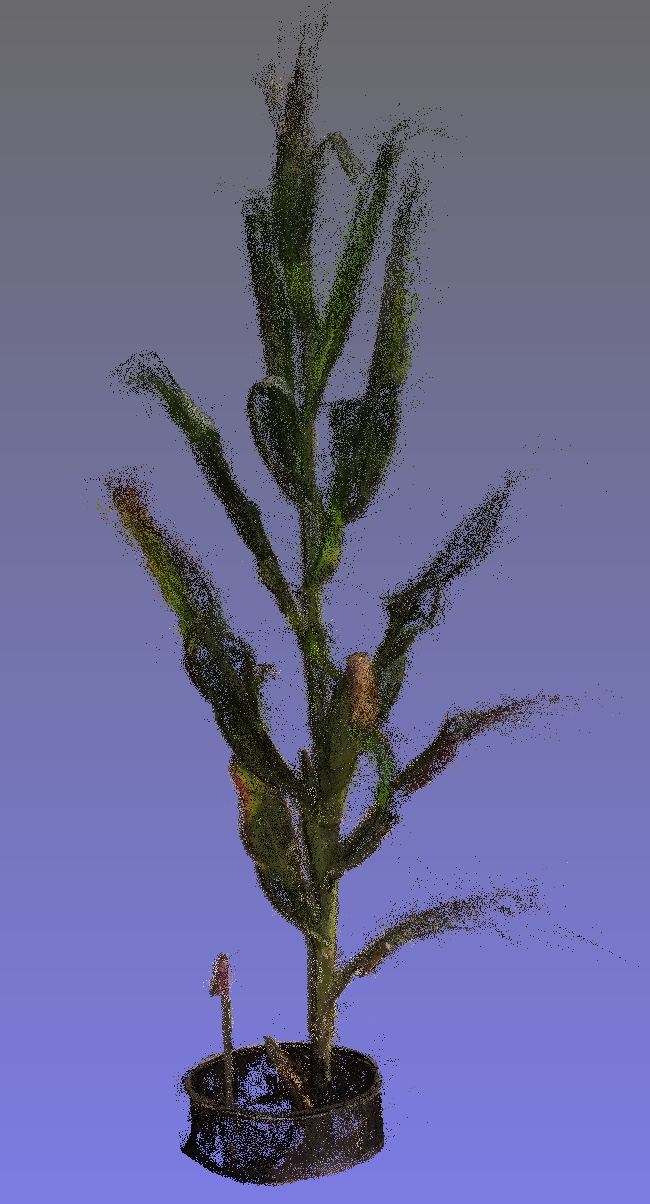}}
{(b) Single orbit, 295 frames; 520,601 points.}
\hfill
\panel{0.287\textwidth}
{\includegraphics[width=\linewidth]{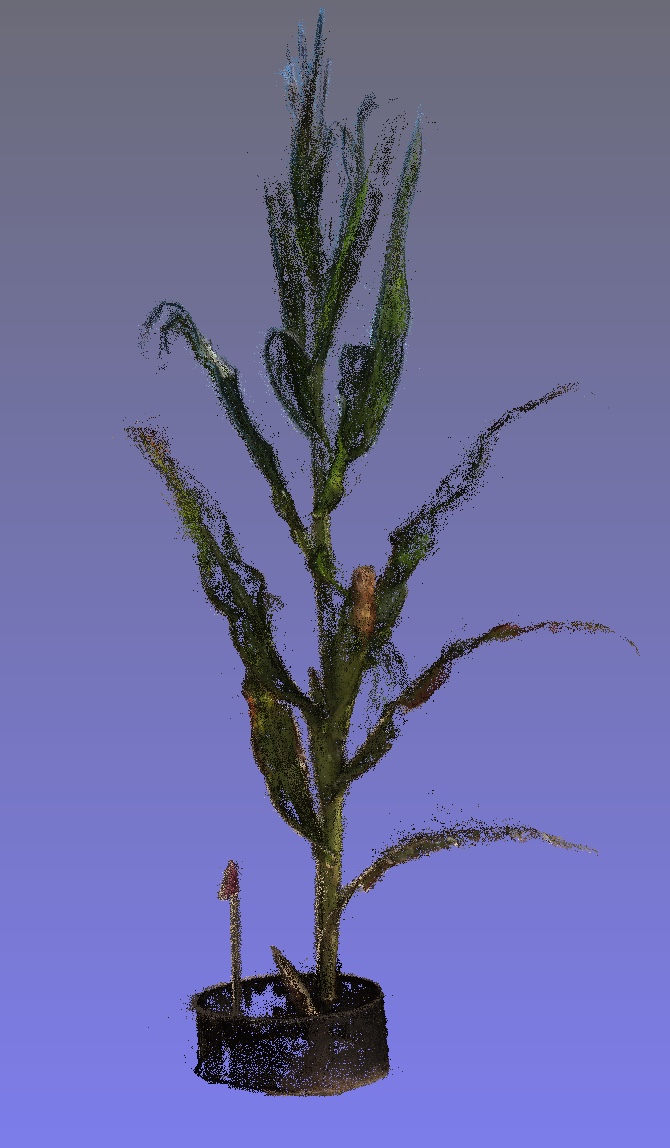}}
{(c) Ribbon skirt, 295 frames; 375,318 points.}

\caption{Comparison of two acquisition configurations
using an equal number of frames from the same
\uav video.
(a) Single-orbit trajectory at approximately
plant height and ribbon skirt trajectory
consisting of three orbits at different
elevations and camera orientations.
(b) The single-orbit reconstruction contains
more points and fewer visible gaps within
the plant structure, but also shows
more stray points outside the plant.
Brown points along some plant boundaries
are consistent with the soil background
visible from the downward camera orientation.
(c) The ribbon skirt reconstruction contains
fewer scattered points outside the plant,
but shows more gaps within the reconstructed
plant structure.
Blue points near some plant boundaries
are consistent with the sky background
visible from approximately parallel viewpoints.
Both reconstructions used 295 frames.
Ground and soil points were removed manually
using comparable procedures in MeshLab,
while above-ground scattered points
were left unedited.
The displayed point counts correspond
to the clouds after ground removal.}
\label{fig:equalframes}
\end{figure}

The equal-frame experiment examined how different
viewpoint distributions and temporal sampling rates
affected the reconstruction of an individual maize
plant while keeping the total number of input frames
constant.
Both reconstructions used 295 frames extracted from
the same source video.
The single-orbit configuration sampled frames at
5 frames/second from one orbital trajectory, with
the camera positioned at approximately the same
height as the plant and oriented slightly downward.
In contrast, the ribbon skirt configuration sampled
frames at 2 frames/second across three consecutive
orbits at different altitudes and camera orientations
(Figure~\ref{fig:equalframes}a).
After comparable manual removal of ground and soil
points, the single-orbit reconstruction contained
520,601 points, while the ribbon skirt reconstruction
contained 375,318 points.
No scattered points above the ground were selectively
removed from either reconstruction.

The two configurations produced visibly different
point-cloud structures
(Figure~\ref{fig:equalframes}b,c).
The single-orbit reconstruction contained more points
within the plant structure and showed fewer visible
gaps in the leaves and other reconstructed organs.
However, it also contained more stray points
distributed outside the recognizable plant structure,
particularly around the leaf boundaries.
In contrast, the ribbon skirt reconstruction showed
fewer stray points and a more clearly defined
distribution of points around the plant.
Nevertheless, several regions of the plant contained
visible gaps, indicating incomplete reconstruction
of some organ surfaces.
Thus, the single-orbit configuration produced a
visually more complete representation of the plant
structure, but with greater point dispersion,
whereas the ribbon skirt produced a cleaner but
less complete point cloud.

Differences in camera orientation were also reflected
in the colors of points near the plant boundaries.
In the single-orbit reconstruction, some points
along the leaf edges appeared brown, consistent
with the soil background visible from the
slightly downward camera orientation.
In the ribbon skirt reconstruction, blue points
were visible along portions of the plant,
particularly in regions observed from the
approximately parallel camera orientation,
where the sky formed part of the background.
These observations suggest that background colors
can be incorporated into reconstructed points
near thin plant boundaries, depending on the
camera viewpoint and surrounding scene.

The quantitative analysis further characterized
differences in point concentration and local
isolation between the two reconstructions
(Table~\ref{tab:scatter}).
The pooled median nearest-neighbor distance
was $s=0.001279$ reconstructed coordinate units
and was used as a common reference spacing
for both point clouds.
The single-orbit reconstruction had a median
nearest-neighbor distance of 0.001304, compared
with 0.001243 for the ribbon skirt reconstruction.
Although the ribbon skirt had a slightly smaller
median nearest-neighbor distance, the single-orbit
cloud had greater median neighborhood counts
at larger common radii.
At $r=5s$, the median neighbor counts were
71 and 66 for the single orbit and ribbon skirt,
respectively.
At $r=10s$, the corresponding counts were
401 and 339.
These results indicate greater local point
concentration at the tested larger radii
in the single-orbit reconstruction.

The ribbon skirt reconstruction, however, had
a slightly smaller proportion of locally
isolated points at each tested threshold.
At $r=3s$, 2.696\% of single-orbit points
and 2.411\% of ribbon skirt points had no
other point within the radius.
At $r=5s$, 1.936\% and 1.678\% of points,
respectively, had at most two neighbors.
Similarly, at $r=10s$, 0.515\% of
single-orbit points and 0.419\% of ribbon
skirt points had at most five neighbors.
These differences are consistent with the
visual observation that the ribbon skirt
contained fewer scattered points outside
the plant structure, although local
isolation does not necessarily indicate
that a point is an outlier.
However, these differences should be interpreted in the context of the unequal point counts.
The single-orbit reconstruction contained approximately 39\% more points than the ribbon skirt reconstruction, which may partly explain the greater number of locally isolated points.
Although the isolation metrics are expressed as proportions rather than absolute counts, differences in the total number and spatial distribution of reconstructed points can affect the proportion falling below each neighborhood threshold.
Therefore, the slightly higher isolation percentages in the single-orbit reconstruction should not be interpreted as evidence that its acquisition strategy inherently produces more outliers.
However, the isolation metrics do not
measure the completeness of reconstructed
plant organs and therefore do not capture
the visible gaps observed in the ribbon
skirt reconstruction.

Overall, the equal-frame comparison revealed
a trade-off between structural coverage
and point dispersion.
The single-orbit configuration retained
more points and showed fewer visible gaps
within the plant structure, but also
introduced more stray points.
The ribbon skirt configuration produced
fewer scattered points outside the plant
but left more regions of the plant
incompletely reconstructed.
Because the two configurations differed
in both temporal sampling rate and
viewpoint distribution, the observed
differences cannot be attributed to
either factor independently.
The results show that an equal number
of input frames does not necessarily
produce equivalent reconstruction
coverage or point-cloud characteristics.
%


\begin{table}
\centering
\scriptsize
\begin{tabular}{lrr}
\toprule
Metric & One orbit, 5 fps & Ribbon skirt, 2 fps\\
\midrule
Retained points & 520,601 & 375,318\\
Bounding-box diagonal & 1.895069 & 1.857731\\
Median nearest-neighbor distance & 0.001304 & 0.001243\\
Median $N_i(5s)$ & 71 & 66\\
Median $N_i(10s)$ & 401 & 339\\
$N_i(3s)=0$ & 2.696\% & 2.411\%\\
$N_i(5s)\leq 2$ & 1.936\% & 1.678\%\\
$N_i(10s)\leq 5$ & 0.515\% & 0.419\%\\
\bottomrule
\end{tabular}
\caption{Point-density and isolation diagnostics for
the equal-frame reconstructions.
Both configurations used 295 frames from the same
source video.
The common reference spacing,
$s=0.001279$ reconstructed coordinate units,
was calculated as the median of the pooled
nearest-neighbor distances from both point clouds.
$N_i(r)$ represents the number of neighboring
points within radius $r$, excluding the query
point itself.
The metrics characterize local point concentration
and isolation but do not directly measure
the completeness or geometric accuracy of
reconstructed plant organs.}
\label{tab:scatter}
\end{table}

\subsection{Scaling from Isolated Plants to Nursery Rows}
\label{sec:scaling}

\begin{figure}[!htbp]
\centering
\panel{0.29\textwidth}
      {\includegraphics[width=\linewidth]{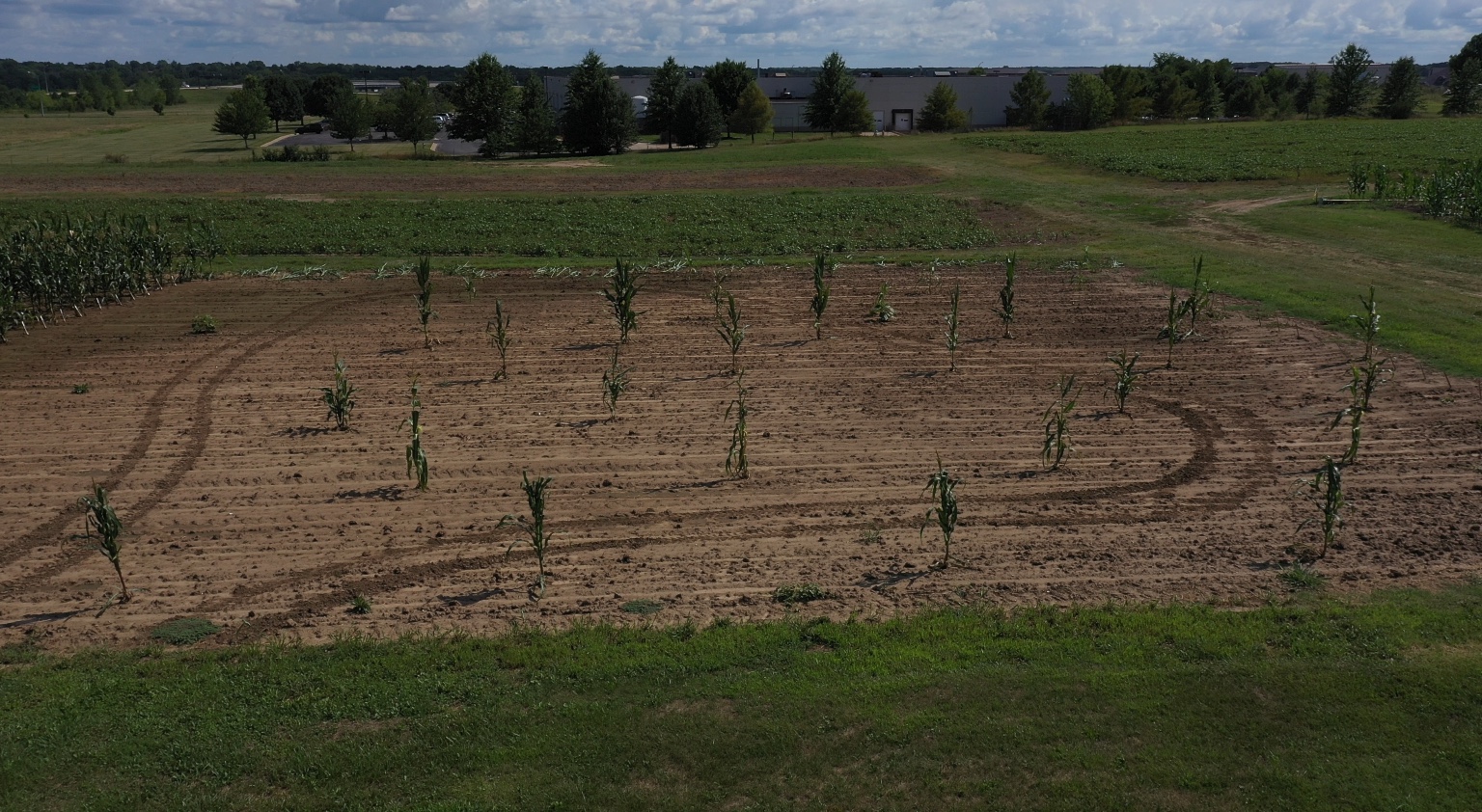}}
      {(a) Representative \rgb frame.}
\panel{0.30\textwidth}
      {\includegraphics[width=\linewidth]{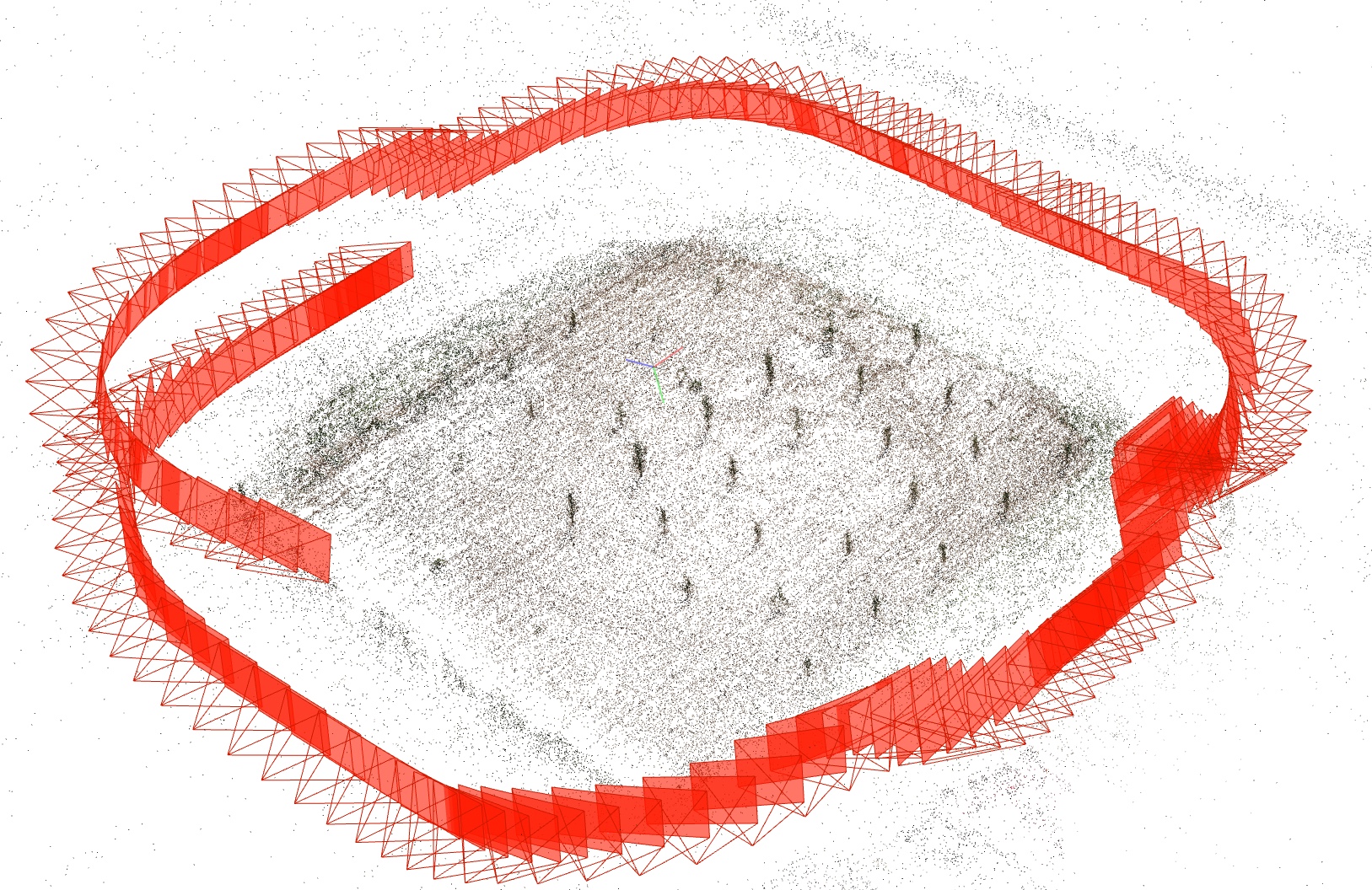}}
      {(b) Estimated camera poses and sparse point cloud.}
\panel{0.39\textwidth}
      {\includegraphics[width=\linewidth]{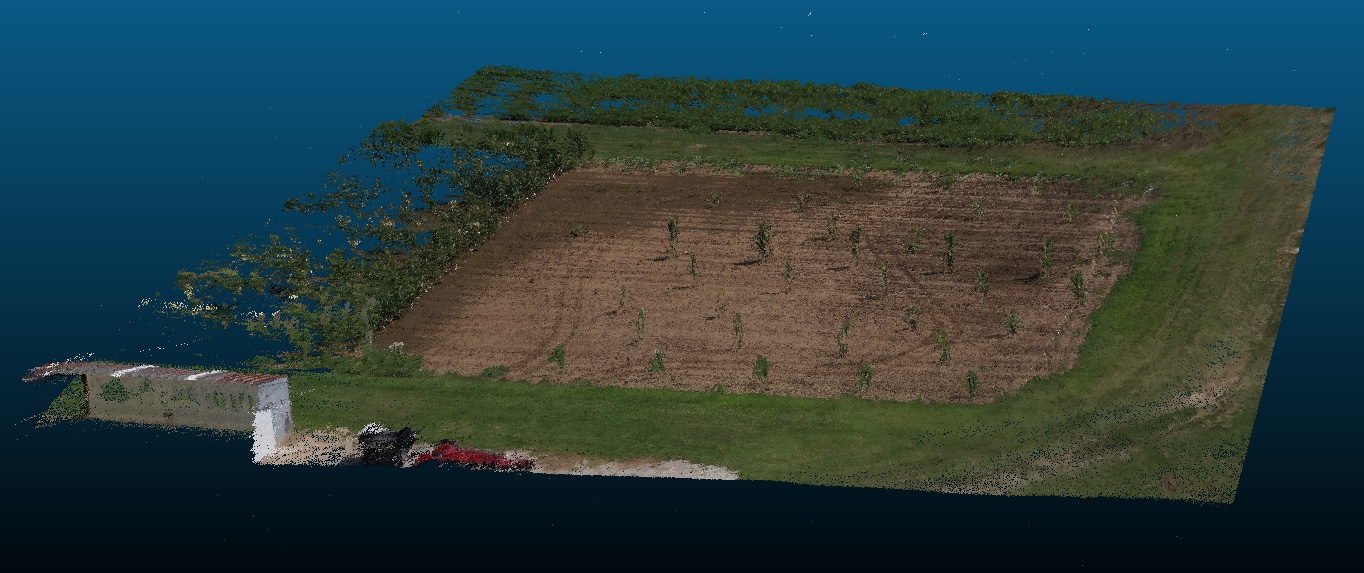}}
      {(c) Dense point cloud.}

\caption{Three-dimensional reconstruction of a
separated plant tiling using one circumferential
orbit with an oblique camera orientation at
approximately 5--7~m altitude.
(a) Representative \rgb frame showing several
spatially separated maize plants.
(b) Estimated camera poses and sparse point cloud.
(c) Dense point cloud showing recognizable
individual plants and surrounding field features,
including tire tracks and the edge of a
neighboring genetic nursery.}
\label{fig:tiling}
\end{figure}

The separated plant tiling provided an intermediate
scenario between reconstructing an isolated maize
plant and a densely planted nursery range.
A circumferential orbit acquired at an approximate
altitude of 5--7~m with an oblique camera orientation
captured several spatially separated plants within
the same scene (Figure~\ref{fig:tiling}a,b).
The initial dense reconstruction contained
20,105,936 points, of which 15,986,041 remained
after cropping.
Individual plants remained recognizable in the
resulting point cloud, although finer morphological
details were less apparent than in the close views
of isolated plants (Figure~\ref{fig:tiling}c).
The separation between plants reduced mutual
occlusion, allowing their overall structures
to remain distinguishable despite the greater
viewing distance.
The reconstruction also captured surrounding
field features, including tire tracks and
the edge of the neighboring genetic nursery.

\begin{figure}[t!]
\centering
\panel{0.35\textwidth}
      {\includegraphics[width=\linewidth]{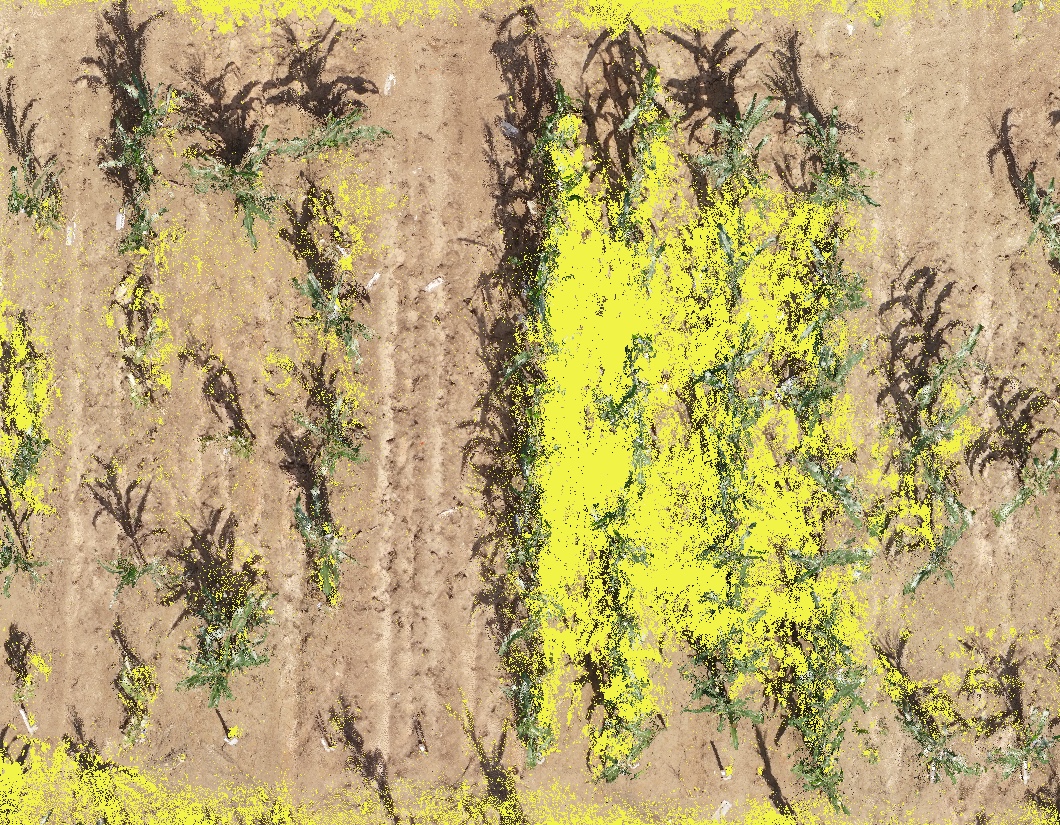}}
      {(a) Single orbit, 126 frames at 1 fps.}
\panel{0.358\textwidth}
      {\includegraphics[width=\linewidth]{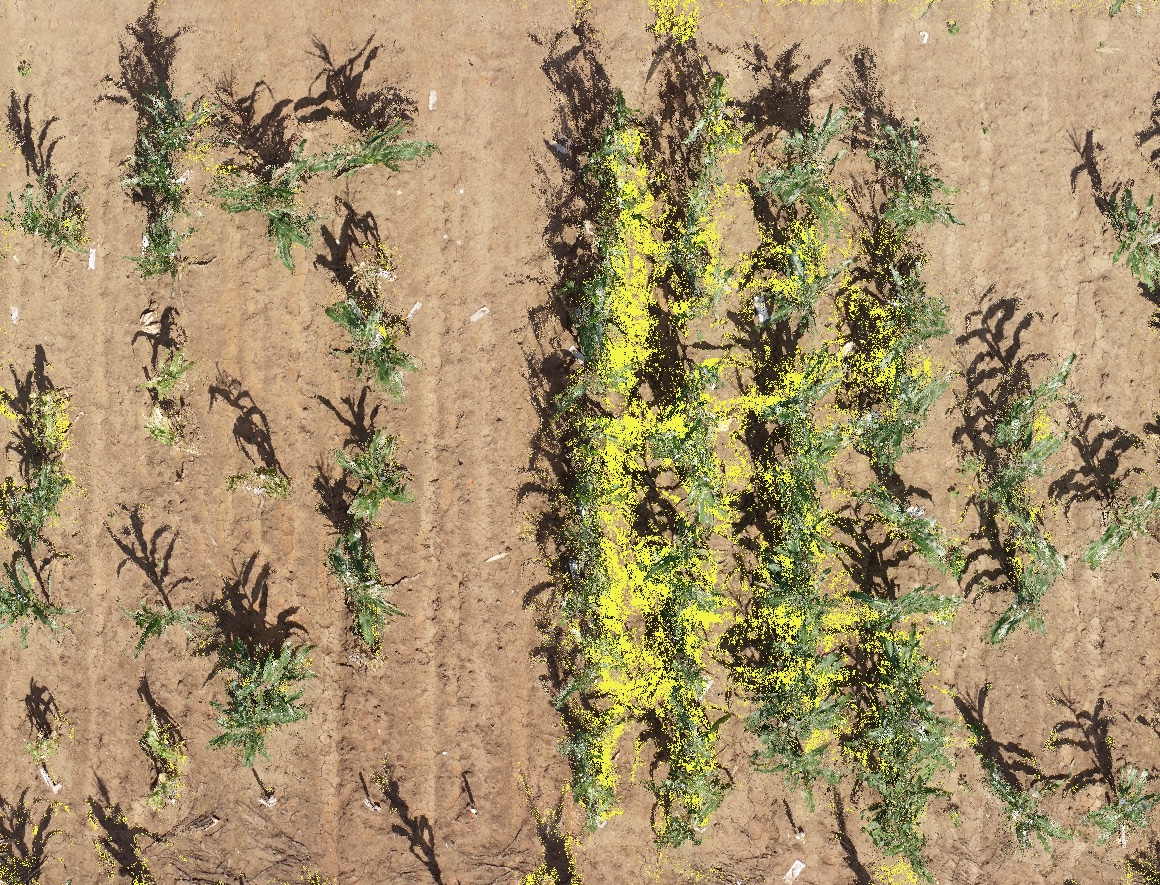}}
      {(b) Orbit and serpentine, 208 frames at 1 fps.}

\panel{0.35\textwidth}
      {\includegraphics[width=\linewidth]{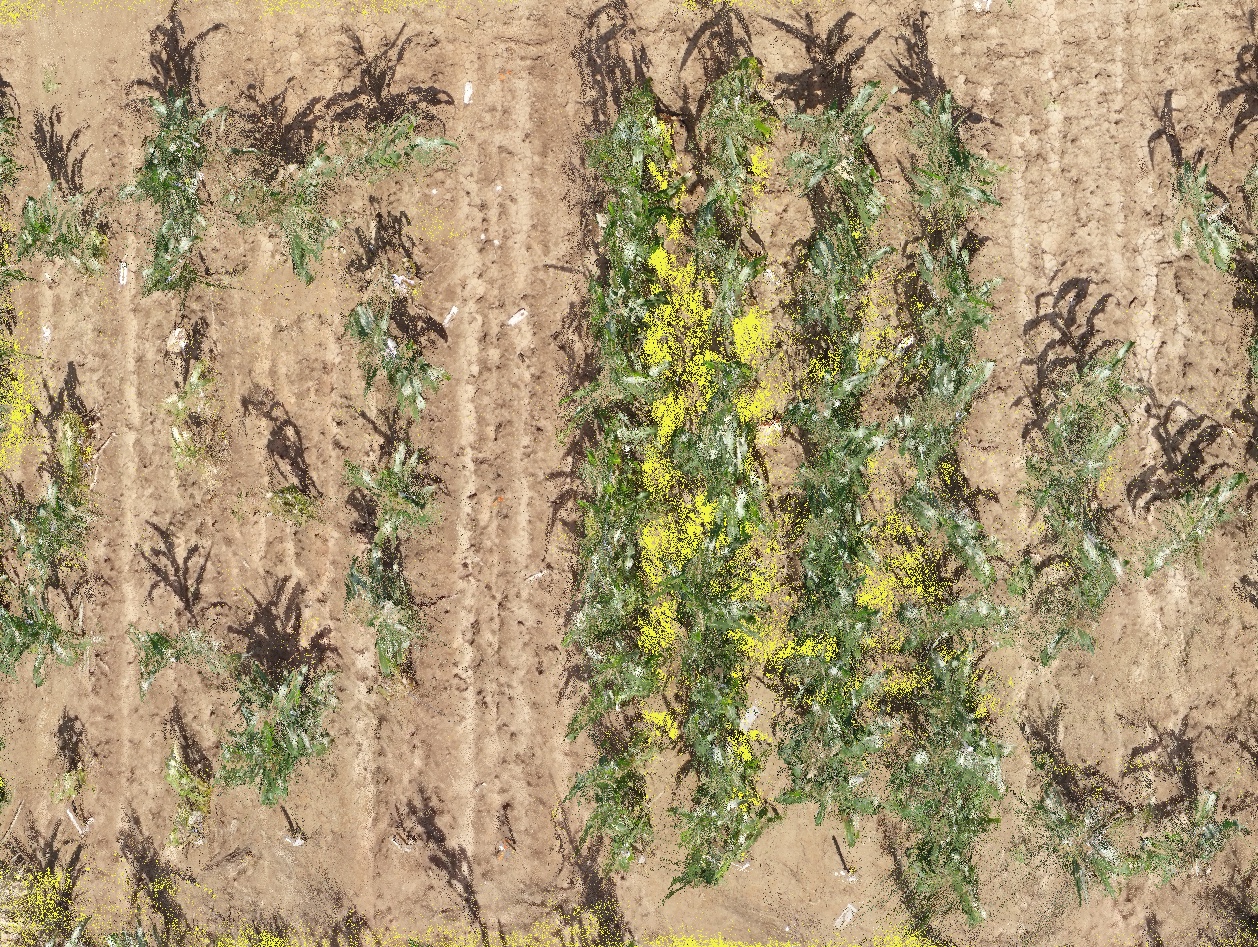}}
      {(c) Single orbit, 426 frames at 3 fps.}
\panel{0.33\textwidth}
      {\includegraphics[width=\linewidth]{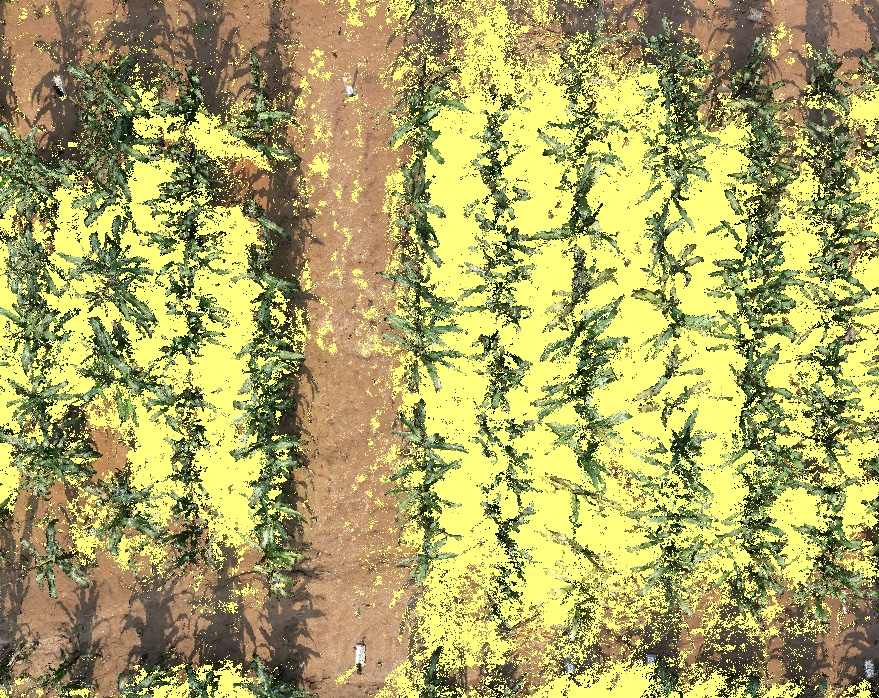}}
      {(d) Curlicue, 287 frames at approximately 0.5 fps.}

\caption{
Close views of four reconstruction experiments conducted on densely planted maize nursery ranges.
Panels (a)--(c) show reconstructions from the same nursery range in 2022, while panel (d) shows the curlicue reconstruction from a different field with the same planting layout in 2024.
(a) A single orbital trajectory with an
oblique camera orientation produced
16,248,202 points from 126 frames.
Substantial voids are visible between
densely planted rows.
(b) Combining orbital and serpentine
trajectories produced 25,369,115 points
from 208 frames.
Additional viewpoints reduced the visible
voids, but leaf movement between passes
produced less sharply defined structures.
(c) Increasing the sampling rate of the
original orbital trajectory to 3 frames/second
produced 77,361,305 points from 426 frames.
Many visible voids were reduced, although
greater point dispersion appeared around
the plant structures.
(d) The curlicue trajectory produced
7,811,270 points from 287 frames sampled
at approximately 0.5 frames/second.
Individual plants appeared more distinguishable,
but substantial reconstruction voids remained.}
\label{fig:range}
\end{figure}

We next examined reconstruction of a densely
planted nursery range containing multiple maize
rows.
Unlike the separated plant tiling, the nursery
contained plants with overlapping leaves,
uneven plant densities, and substantial
occlusion between neighboring rows.
Three reconstruction experiments were conducted
on the same nursery range to examine the effects
of acquisition trajectory and frame sampling
on the resulting point clouds
(Figure~\ref{fig:range}).

The first experiment used a single orbital
trajectory with an oblique camera orientation
at approximately 5--7~m altitude.
A total of 126 frames sampled at 1 frame/second
produced a dense reconstruction containing
16,248,202 points.
The overall arrangement of the nursery rows
and individual plants remained recognizable.
However, the close view revealed substantial
voids in the reconstruction, particularly
between densely planted rows where leaves
overlapped and obscured one another
(Figure~\ref{fig:range}a).
These regions appeared as gaps or bluish
areas in the reconstructed point cloud,
indicating that portions of the plant
structure were not recovered.

In the second experiment, a serpentine
trajectory was added to the orbital acquisition
to provide additional observations of the
nursery range.
The serpentine acquisition included passes
with nadir and oblique camera orientations,
with the oblique view directed perpendicular
to the rows.
The reconstruction used 208 frames sampled
at 1 frame/second and produced
25,369,115 points.
Compared with the single orbit, the additional
viewpoints reduced the visible voids and
recovered more of the plant structure
(Figure~\ref{fig:range}b).
However, the reconstructed leaves appeared
less sharply defined.
Movement between acquisition passes caused
slight differences in leaf positions,
resulting in more dispersed points around
some leaf surfaces.
Thus, the additional trajectory improved
the visible coverage of the nursery plants,
but also introduced artifacts associated
with plant movement.

The third experiment retained the original
orbital trajectory but increased the frame
sampling rate from 1 to 3 frames/second,
resulting in 426 input frames.
The resulting reconstruction showed fewer
visible voids than the lower sampling rate
reconstructions, with many previously
incomplete regions becoming more densely
represented (Figure~\ref{fig:range}c).
However, the increased number of frames
also produced greater point dispersion
around the reconstructed plant structures.
This dispersion was consistent with leaf
movement captured across the additional
frames, which introduced differences
between observations of the same organs.

The fourth experiment used a curlicue trajectory in a different maize field during the 2024 growing season.
The trajectory was intended to provide more varied viewpoints of densely planted maize and examine whether this acquisition pattern could reduce reconstruction artifacts.
A total of 287 frames were extracted at approximately 0.5 frames/second from a video lasting 9 minutes and 33 seconds.
The resulting dense reconstruction contained 7,811,270 points, fewer than the three reconstructions from the 2022 nursery range.
Despite the smaller point cloud, individual plants were clearly distinguishable from their neighbors, without visible duplicated or ghost plant structures or double-layered soil surfaces (Figure~\ref{fig:range}d; Figure~\ref{fig:curlicue_detail}).
However, substantial voids remained, primarily on the reconstructed soil surface.
Thus, the curlicue acquisition produced recognizable individual plant structures without visible plant duplication, although the surrounding ground surface remained incompletely reconstructed.
%
%
%

%
%

The quantitative evaluation of visible
reconstruction voids further characterized
the differences among the nursery range
reconstructions.
In the 2022 field experiment, voids occupied
17.57\% of the rendered scene from the
single orbit sampled at 1 frame/second.
The visible void area decreased to 5.07\%
with the addition of serpentine passes
and to 2.68\% when the original orbit
was sampled at 3 frames/second.
In the 2024 curlicue experiment, visible
voids occupied 37.50\% of the selected
scene, primarily on the soil surface.
Although the curlicue reconstruction
contained a larger proportion of visible
voids, individual plants remained
distinguishable without visible duplicated
structures.
%
%
Because the curlicue experiment was
conducted in a different field and
growing season, its point counts and
void percentage
are not directly comparable with those
of the 2022 reconstructions.
The nursery range experiments revealed different reconstruction characteristics associated with acquisition trajectory and frame sampling.
In the 2022 experiments, the single orbit preserved the overall nursery arrangement but contained substantial gaps in regions with overlapping vegetation.
Adding serpentine passes reduced these gaps while introducing greater point dispersion around some leaf surfaces.
Increasing the frame sampling rate along the original orbit further reduced visible voids, but also increased point dispersion.
The curlicue trajectory, evaluated in a different field during the 2024 growing season, produced a smaller point cloud in which individual plants were clearly distinguishable without visible duplicated or ghost plant structures.
However, substantial reconstruction voids remained, primarily on the soil surface.
These observations suggest that reducing visible plant duplication and improving the separation of neighboring plants do not necessarily eliminate gaps in the reconstructed scene.
The curlicue result provides an additional example of reconstruction behavior rather than a controlled comparison with the three earlier acquisition strategies.

\subsection{Failure Modes and Limits of Postprocessing}
\label{sec:failure_modes}

The reconstruction experiments revealed several
limitations that became particularly apparent
when imaging densely planted nursery ranges.
Occlusion and shadowing were associated with
missing regions in both plant and soil surfaces.
In the nursery reconstructions, these gaps
were most visible where neighboring plants
overlapped or where portions of the scene
were poorly observed from the available
camera viewpoints.
Although additional viewpoints and frames
reduced some of these gaps
(Section~\ref{sec:scaling}), they did not
eliminate all reconstruction artifacts.

\begin{figure}[t!]
\centering
\panel{0.47\textwidth}
      {\includegraphics[width=\linewidth]{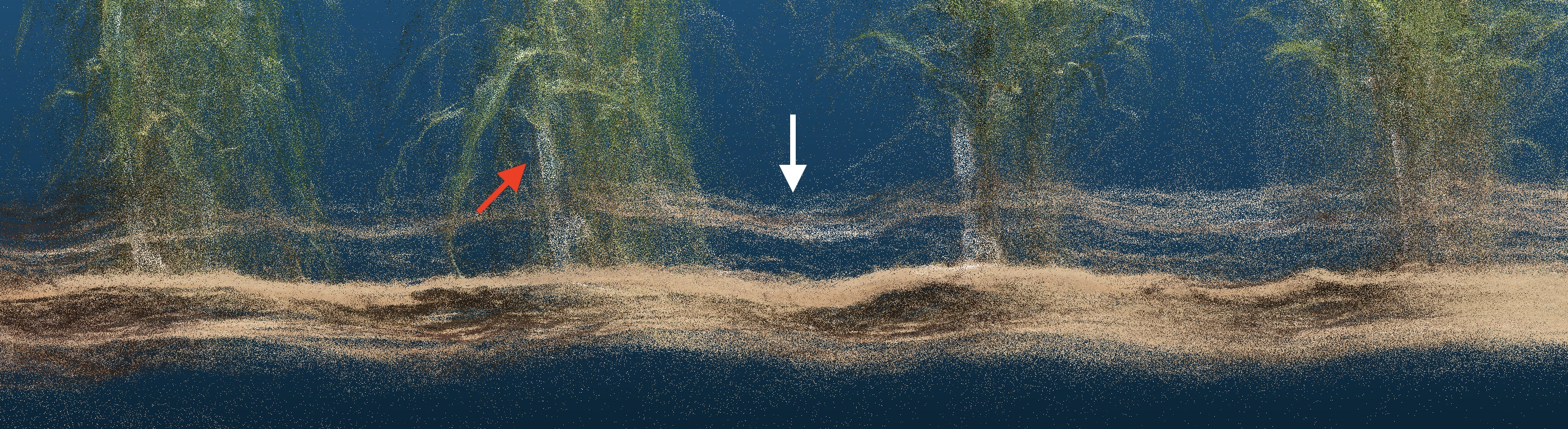}}
      {(a) Misregistered soil surfaces.}
\hfill
\panel{0.47\textwidth}
      {\includegraphics[width=\linewidth]{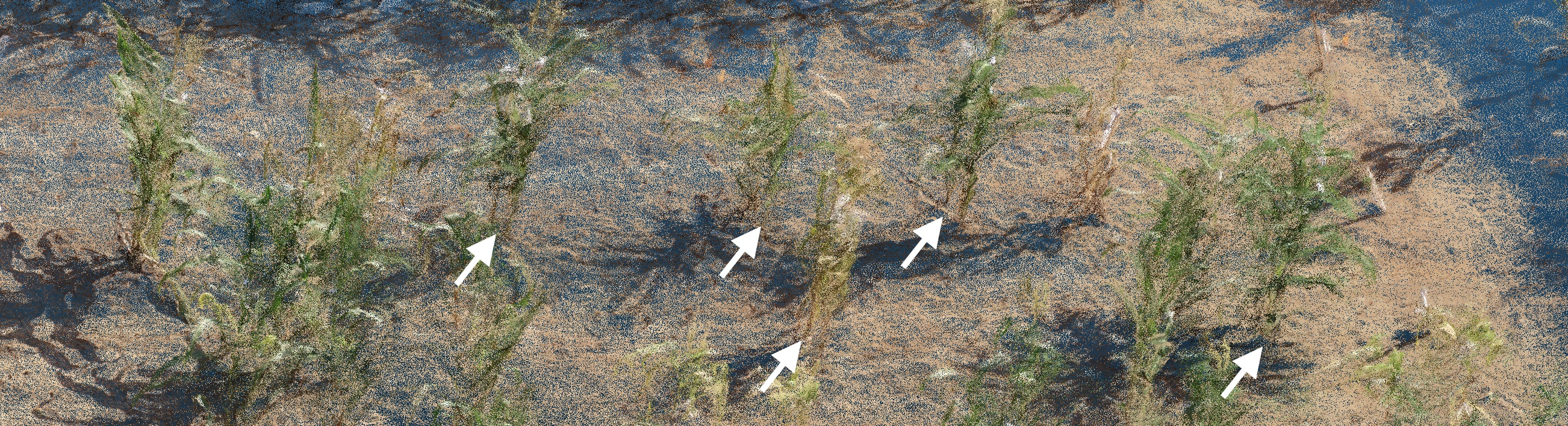}}
      {(b) Duplicated plant structures.}

\caption{Examples of geometric reconstruction
artifacts observed in densely planted maize
nursery ranges.
(a) Separated soil surfaces indicating
inconsistent reconstruction of the ground.
(b) Duplicated plant structures that do not
correspond to distinct physical plants.
These artifacts illustrate errors in
scene reconstruction and registration
that are separate from gaps caused by
occlusion or insufficient observations.}
\label{fig:artifacts}
\end{figure}

\begin{figure}[b!]
\centering
\panel{0.48\textwidth}
      {\includegraphics[width=\linewidth]{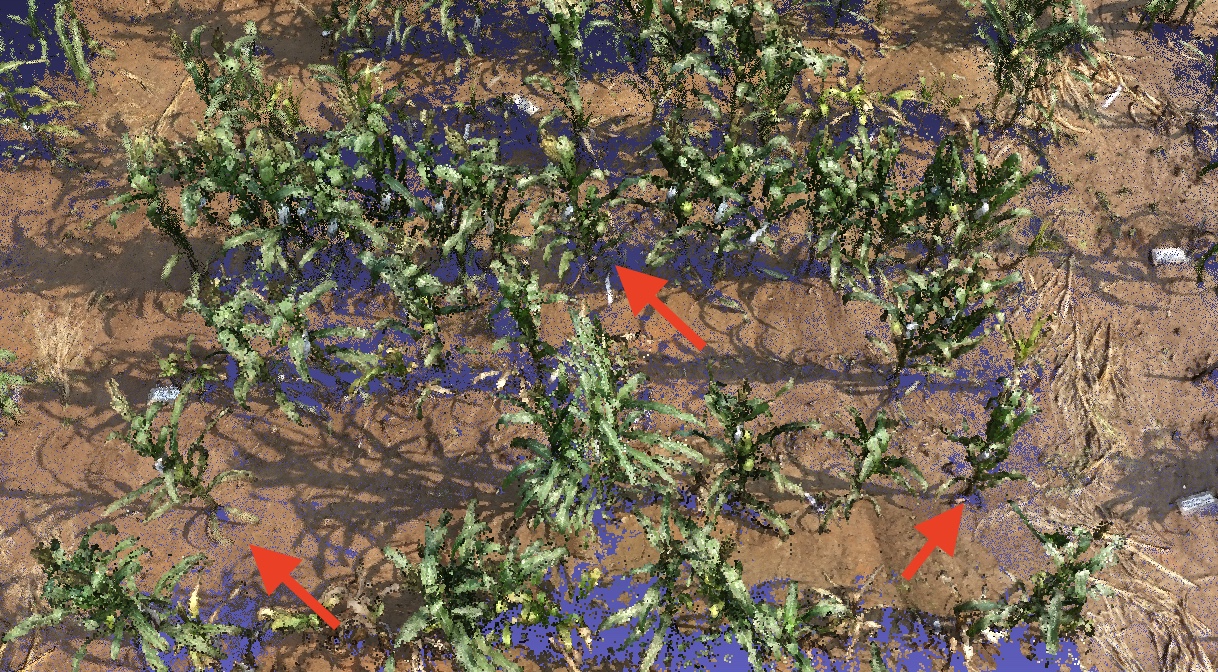}}
      {(a) Individual plants.}
\panel{0.48\textwidth}
      {\includegraphics[width=\linewidth]{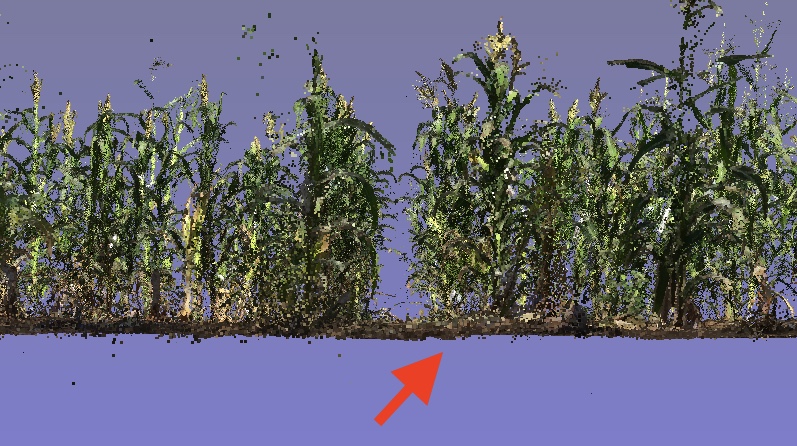}}
      {(b) Soil surface.}

\caption{Detailed views of the 2024 curlicue reconstruction.
(a) Individual maize plants are distinguishable without visible
duplicated or ghost plant structures. Red arrows indicate
representative plants.
(b) The reconstructed soil surface appears as a single layer,
without visible double-layering. The red arrow indicates
the reconstructed soil surface.}
\label{fig:curlicue_detail}
\end{figure}

Other failures involved the placement of
reconstructed structures rather than
missing points.
In some nursery reconstructions, the soil
appeared as separated or overlapping planes,
while individual plants were duplicated
(Figure~\ref{fig:artifacts}a,b).
These structures did not correspond to
the physical arrangement of the nursery
and indicated inconsistencies in the
reconstruction and registration of the scene.
The repetitive appearance of maize plants,
similar leaf structures, and changes in
plant position between observations
presented additional challenges for
matching corresponding image features.
However, the individual contributions
of these factors were not evaluated
separately.
In contrast, the curlicue reconstruction from the 2024 growing season
showed no visible duplicated plant structures or double-layered soil
surfaces in the inspected regions (Figure~\ref{fig:curlicue_detail}).
Red arrows indicate representative individual plants and the
single-layered soil surface.
Although these artifacts were not observed in the selected views,
substantial reconstruction voids remained, particularly on the soil
surface.

Plant movement also affected reconstruction
at the individual plant scale.
As observed in Section~\ref{sec:individual},
moving leaves could produce ghost organs
or dispersed points around leaf margins,
tips, and tassels.
Ambient wind and \uav rotor downwash
were possible contributors to this movement,
although their individual effects were
not measured.
Unlike gaps caused by missing observations,
these artifacts introduced conflicting
representations of the same plant structure.
Consequently, the reconstructed point
cloud could contain both missing and
duplicated geometry.

Color masking was evaluated as a way to
reduce background content before
three-dimensional reconstruction.
However, the color characteristics of
maize organs overlapped with those of
the surrounding environment.
The \exg and \emph{L*a*b*} masks could
suppress soil and other background
regions, but also removed genuine
plant tissue with brown, yellow,
orange, or red coloration.
This was particularly problematic for
senescent leaves, tassels, and other
structures that were not predominantly
green.

\begin{figure}[t!]
\centering
\panel{0.25\textwidth}
      {\includegraphics[width=\linewidth]{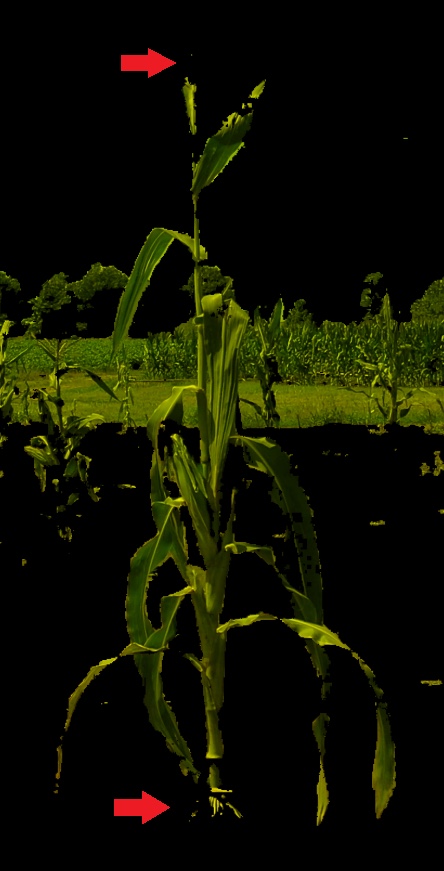}}
      {(a) \exg-masked \rgb frame.}
\hfill
\panel{0.25\textwidth}
      {\includegraphics[width=\linewidth]{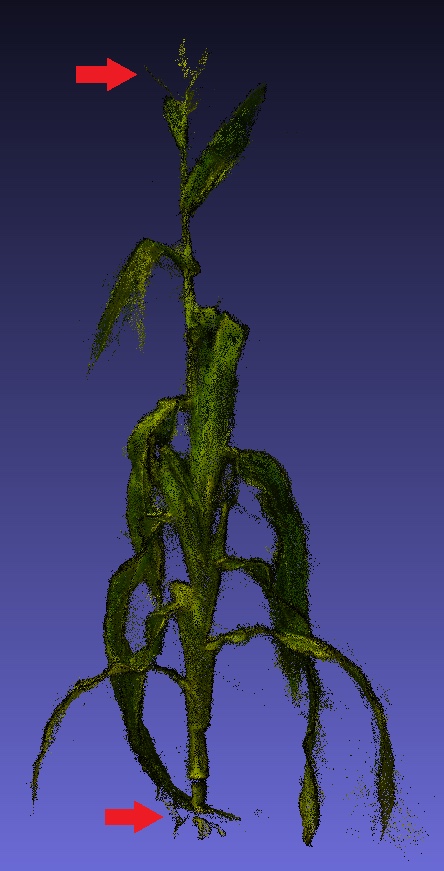}}
      {(b) Reconstructed point cloud.}
\hfill
\panel{0.239\textwidth}
      {\includegraphics[width=\linewidth]{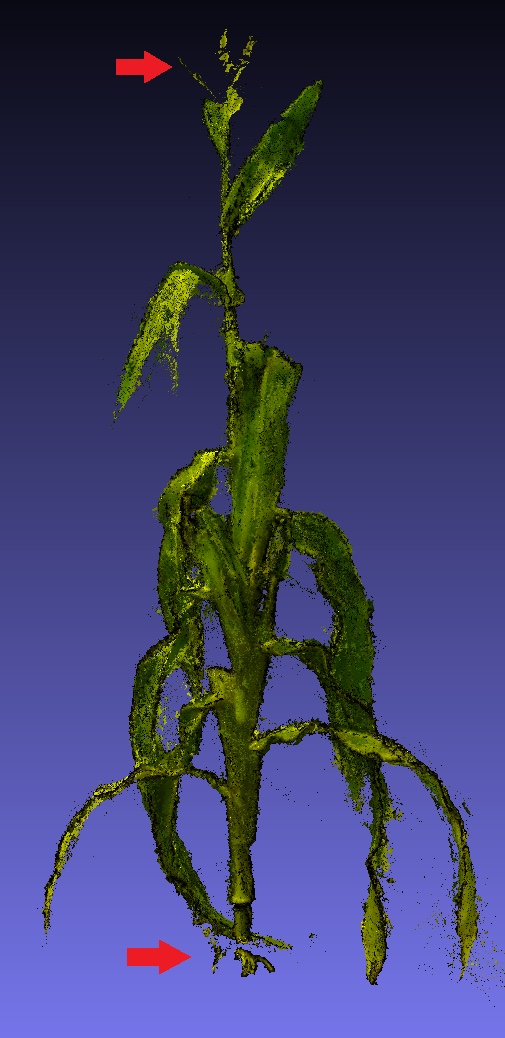}}
      {(c) Ball pivoting surface mesh.}

\caption{Effects of \exg masking on the
reconstruction of an individual maize plant.
(a) Representative \rgb frame after
applying the \exg mask.
(b) Reconstructed point cloud showing
missing plant structures following
background suppression.
The upper red arrow identifies the
missing tassel, while the lower
red arrow identifies missing brace roots.
These structures were excluded because
their colors did not satisfy the
green vegetation mask.
(c) Triangular surface mesh generated
using ball pivoting.
The missing structures remain absent
from the surface reconstruction because
the corresponding image information
was removed before reconstruction.}
\label{fig:masking}
\end{figure}

The effect was visible in the reconstruction
generated from \exg-masked images
(Figure~\ref{fig:masking}).
The initial point cloud contained
2,842,193 points, which decreased to
463,607 after cropping to the plant
region and 378,646 after removing
black boundary points.
These reductions reflected successive
processing operations rather than
improvements in reconstruction accuracy.
Although masking suppressed portions
of the background, it also removed
image information needed to reconstruct
genuine plant structures.
The resulting point cloud lacked
portions of the tassel and brace roots,
as indicated by the red arrows in
Figure~\ref{fig:masking}.

Applying the ball pivoting algorithm  to generate a triangular surface mesh did not restore
the missing structures 
(Figure~\ref{fig:masking}c).
Ball pivoting connected available
points into surface faces, but could
not reconstruct organs whose image
information had been removed by masking.
The resulting surface therefore retained
gaps corresponding to missing plant
structures.
This example illustrates a limitation
of color based background suppression:
reducing unwanted scene content can
also eliminate biologically meaningful
structures when plant and background
colors overlap.

%
%
%
%
%
%

%
%
%

\section{Discussion}

\subsection{From Visual Reconstruction to Quantitative Plant Phenotyping}

The reconstruction of individual maize plants demonstrates that consumer \uav \rgb imagery can preserve substantial morphological information under field conditions without requiring a fixed imaging platform or controlled background. 
The reconstructed point clouds retained recognizable leaves, stems, brace roots, tassel branches, ears, and silks. 
These observations are important because the geometry of these organs provides information that is difficult to recover from conventional two-dimensional images alone. 
For example, leaf orientation, the spatial arrangement of leaves around the stem, and the position of reproductive structures are inherently three-dimensional characteristics.  
However, recognizing an organ in a point cloud is different from establishing that its dimensions or geometry have been reconstructed accurately.  
The present results demonstrate the visibility of these structures, but do not yet establish the accuracy required for quantitative phenotyping.
Previous studies have demonstrated the potential of three-dimensional plant reconstruction for extracting morphological traits under more controlled acquisition conditions.  
\citet{li2022c} reconstructed maize seedlings using multiple views and reported strong correlations between reconstructed traits and manual measurements, including plant height, leaf length, leaf width, and relative leaf area.  
They further demonstrated 
organ segmentation for extracting structural traits from maize point clouds \citep{li2022d}, while \citet{yang2024} investigated stem and leaf segmentation across a dataset of 428 maize plants.  
These studies establish that three-dimensional representations can support quantitative measurements when the reconstructed geometry is sufficiently reliable.  
Our experiments address a different part of this problem: whether recognizable plant structures can be recovered from freely acquired \uav imagery in an agricultural field, where the plants are not stationary, the background is uncontrolled, and neighboring plants may obscure one another.
This distinction becomes particularly important when moving from isolated plants to nursery rows.  
At the individual plant scale, the camera could observe the plant from multiple directions with relatively little interference from neighboring vegetation.  
As the acquisition expanded to separated plants and densely planted nursery ranges, the reconstruction retained the general plant arrangement but progressively lost the continuity and separation of individual organs.  
Similar challenges have been considered by \citet{zermas2020}, who reconstructed small groups of maize plants from \uav and handheld \rgb imagery and extracted traits including plant count, height, leaf area index, leaf length, and leaf angle.  
Our results support the feasibility of reconstructing maize from aerial \rgb imagery, but also show why extending organ measurements to dense nursery populations remains challenging.  
A point cloud may represent the overall arrangement of a nursery while still lacking the geometry needed to distinguish individual leaves or measure their dimensions.
This difference suggests that the usefulness of a reconstruction should be evaluated according to the intended phenotype rather than through a single measure of point cloud quality.  
A reconstruction containing incomplete leaf surfaces may still support plant localization, row identification, or approximate canopy measurements, whereas estimating leaf area, leaf curvature, or the angle between a leaf and the stem requires more complete and correctly associated organ geometry.  
Similarly, a point cloud may be sufficient for identifying certain structures, while other measurements may require a surface mesh or explicit organ segmentation.  
The appropriate representation therefore depends on the biological question.  
Future phenotyping workflows should distinguish between traits that can be extracted reliably from incomplete point clouds and those that require additional reconstruction or validation.

\subsection{Viewpoint Diversity, Plant Movement, and Reconstruction Completeness}

One of the central findings of this study is that increasing the number of observations does not necessarily produce a better reconstruction in every respect.  
In the equal frame comparison, the single orbit and ribbon skirt configurations used the same number of images but distributed their observations differently.  
The single orbit produced more points and fewer visible gaps within the plant structure, although it also contained more scattered points outside the plant.  
%
%
%
%
%
These observations show that point density, point isolation, and visible structural continuity describe different properties of a reconstruction and should not be treated as interchangeable measures of quality.
The difference can be understood in terms of the information supplied by the camera viewpoints.  
Repeated observations from a similar elevation can provide substantial overlap for surfaces already visible from that trajectory, producing dense point coverage in those regions.  
However, additional frames from similar viewpoints cannot recover surfaces that remain hidden.  
Changing camera elevation or orientation can expose previously concealed structures, but distributing a fixed number of frames across more viewpoints also changes the amount of overlap available within each viewing region.  
Consequently, greater viewpoint diversity does not guarantee a denser or more continuous point cloud.  
The balance between observing new surfaces and maintaining sufficient overlap for reliable correspondence is particularly important for maize, where leaves are thin, curved, and frequently overlap.
These findings are consistent with earlier work demonstrating the importance of viewing geometry in crop reconstruction.  
\citet{che2020} reported that nadir and oblique imagery could support maize height estimation, while oblique observations provided advantages for representing canopy layers in voxel based leaf area index estimation.  
\citet{niu2024a} also reported improvements in maize reconstruction using oblique imagery compared with nadir observations, together with reduced height accuracy at coarser spatial resolution.  
Our experiments extend this discussion to individual organ visibility and the spatial distribution of reconstructed points.
Rather than identifying one camera orientation as universally preferable, the results indicate that different viewpoints contribute different information about the plant.  
Parallel views can reveal vertical plant profiles, while elevated and oblique observations can expose upper leaves, tassels, and surfaces that are poorly visible from lower positions.
The nursery range experiments further illustrate why the relationship between additional observations and reconstruction quality is not straightforward.  
The initial orbital reconstruction preserved the row arrangement but contained substantial voids between overlapping plants.  
Adding serpentine passes reduced these gaps by introducing observations from additional directions.  
Increasing the sampling rate of the original orbit also reduced visible voids.  
However, both approaches produced greater point dispersion around some plant structures.  
The additional observations therefore contributed information about previously incomplete regions while also introducing inconsistencies associated with changes in plant position during acquisition.
The curlicue reconstruction from the 2024 growing season provided a different outcome.
Although substantial voids remained, primarily on the soil surface, individual plants were clearly distinguishable without visible duplicated or ghost plant structures, and the inspected soil surfaces showed no visible double-layering.
The reconstruction contained fewer points than the three earlier nursery range reconstructions, suggesting that a larger point cloud is not necessarily associated with better separation of neighboring plants.
This observation also illustrates that the completeness of the reconstructed ground surface and the ability to distinguish individual plants are different aspects of reconstruction quality.
However, because the curlicue acquisition was conducted in a different field and growing season, the absence of visible plant duplication cannot be attributed to the trajectory alone.
Unlike rigid objects, maize plants do not maintain exactly the same geometry throughout a flight.  
Leaves bend, twist, and move independently, particularly near their tips and in younger upper leaves.  
When the camera returns to a plant after several seconds, the same leaf may occupy a different position.  
Structure from Motion and multiview stereo reconstruction generally assume that corresponding observations describe a static scene.  
When this assumption is violated, the reconstruction may contain duplicated leaf surfaces, ghost organs, or scattered points rather than a single consistent representation.  
The observed artifacts therefore reflect a fundamental difficulty in reconstructing living plants: increasing the number of viewpoints can improve visibility, but acquiring those viewpoints over time also increases the opportunity for the plant to change shape.
This creates a trade-off between spatial coverage and temporal consistency.  
A longer acquisition can observe more of the plant, but the resulting images may no longer describe one fixed plant geometry.  
Conversely, a shorter acquisition may reduce differences in organ position while leaving some surfaces insufficiently observed.  
The optimal strategy is therefore unlikely to depend on frame count alone.  
It should consider camera geometry, overlap, acquisition duration, and the movement of the target plant.  
Future experiments using synchronized cameras or repeated acquisitions under measured wind conditions could help separate the effects of viewpoint diversity from those of temporal movement.  
The present study identifies these competing factors, but does not independently quantify their contributions.

\subsection{Missing Geometry, Reconstruction Artifacts, and the Limits of Postprocessing}

The reconstruction failures observed in this study represent different types of information loss and geometric inconsistency. 
Occlusion produces missing observations when leaves, stems, or soil surfaces are hidden from the available viewpoints.  
Plant movement produces conflicting observations when the same organ occupies different positions across frames.  
Incorrect image correspondence or registration can produce duplicated plants or separated soil surfaces that do not correspond to the physical scene.  
Although these failures may all appear as irregularities in a point cloud, they have different causes and cannot necessarily be corrected using the same processing strategy.
The distinction is especially relevant when interpreting point cloud density.  
A densely reconstructed region is not necessarily geometrically accurate, because some of its points may represent moving organs, background surfaces, or incorrect correspondences.  
Conversely, a region with relatively few points may correspond to a genuine thin structure, such as a tassel branch or leaf tip.  
This limitation is consistent with findings from \uav LiDAR studies, where increasing point density beyond a certain level did not necessarily improve crop height estimation \citep{luo2021a}.  
Although LiDAR and image based reconstruction generate points through different processes, both demonstrate why point count alone is insufficient for evaluating the usefulness of three-dimensional plant data.  
For photogrammetric reconstructions, point density must be considered alongside structural continuity, organ visibility, and geometric consistency.
Postprocessing introduces another challenge because removing unwanted points can also remove biologically meaningful information.  
In our masking experiment, \exg- and \emph{L*a*b*}-based approaches reduced portions of the background but also excluded plant structures that were not predominantly green.  
The missing tassel and brace roots illustrate this problem.  
A green vegetation mask implicitly assumes that the relevant plant structures have a particular color, yet maize organs vary in pigmentation with developmental stage, genotype, environmental conditions, and tissue condition.  
Senescent leaves, reproductive organs, and some phenotypic abnormalities may therefore be removed together with soil or other background material.  
This is particularly problematic for phenotyping studies because the removed structures may be the very traits researchers intend to measure.
The same consideration applies to surface reconstruction.  
Ball pivoting converted the dense point clouds into triangular meshes and made some morphological structures easier to distinguish, including silks, leaf midribs, and tassel branches.  
However, the resulting surfaces were constrained by the available point cloud.  
Missing points could produce holes, while duplicated or displaced points could lead to irregular surface connections.  
In the original reconstruction experiments, MeshLab ball pivoting provided visually finer representations of certain plant structures than the alternative surface reconstruction methods examined.  
Nevertheless, a visually smoother or more continuous surface should not be interpreted as evidence of greater geometric accuracy without an independent reference.  
Surface reconstruction can change how the available geometry is represented, but it cannot establish the correct shape of an organ that was never adequately observed.
These results suggest that preprocessing and reconstruction should be evaluated with respect to the biological information they preserve, not only the background points they remove or the visual appearance of the final model.  
For example, excluding soil may simplify an isolated plant reconstruction, but cropping too close to the ground can eliminate brace roots, which are relevant to plant anchorage and lodging resistance.  
Similarly, filtering scattered points may improve the appearance of a leaf while removing legitimate observations of thin or damaged tissue.  
A phenotyping workflow therefore requires a distinction between artifacts and biological variation.  
This distinction becomes especially important when reconstructing plants with unusual morphology, because an automated procedure designed around typical plant appearance may remove precisely the structures that distinguish an abnormal phenotype.

\subsection{Toward a Public Dataset of Field Grown Plants Reconstructed from RGB UAV Imagery}

Beyond demonstrating the feasibility of reconstructing individual maize plants and nursery ranges, these experiments provide a starting point for developing a publicly accessible dataset of three-dimensional plant reconstructions derived from consumer \uav \rgb imagery.  
Such a dataset would address a different need from collections of images or point clouds acquired under controlled laboratory conditions.  
Field grown plants experience changing illumination, wind, background variation, mutual occlusion, and differences in growth stage and morphology. These conditions complicate reconstruction, but they are also the conditions under which many agricultural phenotyping systems must eventually operate.  
A dataset that preserves these challenges would provide a useful resource for studying the robustness and limitations of reconstruction algorithms.
Existing plant reconstruction and segmentation studies demonstrate the value of structured three-dimensional data for quantitative phenotyping.  
For example, the maize dataset examined by  \citet{yang2024} supported learning stem and leaf segmentation across hundreds of plants, while the studies by \citet{li2022c,li2022d} demonstrated the extraction of morphological information from reconstructed maize.  
These efforts establish the importance of having plant representations that can be segmented and measured.  
A complementary field dataset could focus on the acquisition conditions and reconstruction variability encountered when those representations are generated from freely flown \uav imagery.  
Its contribution would not simply be to provide additional point clouds, but to connect the reconstructed geometry with the images, camera viewpoints, field conditions, and processing decisions that produced it.
The experiments in this study illustrate why retaining the complete acquisition and processing history would be important.  
The equal frame comparison showed that the same number of images can produce point clouds with different structural coverage and point distributions.  
The nursery range experiments showed that additional trajectories or frames can reduce visible voids while introducing greater dispersion around moving leaves.  
The masking experiment demonstrated that preprocessing can remove genuine plant structures before reconstruction begins.  
If only the final point clouds were retained, much of the information needed to understand these differences would be lost.  
A reproducible dataset should therefore include the original \rgb images or video, frame selection information, camera calibration and estimated poses, reconstruction parameters, intermediate point clouds, surface meshes where available, and records of masking, cropping, and manual editing.
A useful dataset should also represent variation in the plants themselves.  
The present experiments include isolated plants, separated plant tilings, and densely planted nursery rows, each presenting different reconstruction challenges.  
Expanding this collection across genotypes, developmental stages, planting densities, and environmental conditions would allow researchers to examine whether reconstruction methods generalize beyond a small number of favorable examples.  
Including plants with senescent leaves, damaged tissue, reproductive structures, or unusual morphology would be particularly valuable.  
Such variation would help determine whether an algorithm preserves biologically meaningful differences or preferentially reconstructs plants that conform to a typical appearance.
An important distinction is needed between a dataset of reconstructed plants and a dataset containing validated three-dimensional ground truth.  
The point clouds generated in this study are reconstruction outputs, not independently verified measurements of the physical plants.  
They could support algorithm development and comparisons of reconstruction behavior, but they should not be treated as exact geometric references. Developing a benchmark for reconstruction accuracy would require an additional validation component, such as calibrated reference objects, independent measurements of selected organs, or higher fidelity reference scans.  
Repeated acquisitions of the same plants could also help quantify reconstruction variability under different flight conditions.  
These measurements would make it possible to distinguish errors caused by acquisition geometry, plant movement, and the reconstruction algorithm itself.
The longer term opportunity is to connect image acquisition, three-dimensional reconstruction, organ segmentation, and biological measurements within a common dataset.  
For example, a reconstructed plant could be associated with annotations identifying stems, leaves, tassels, ears, and brace roots, together with measurements of selected morphological traits.  
Such annotations would support the development of algorithms that identify organs in incomplete point clouds and evaluate which traits can be recovered reliably.  
The original images and camera poses would also allow researchers to test alternative reconstruction methods on the same observations, rather than comparing outputs generated from unrelated datasets.  
This would make it possible to investigate whether improvements arise from acquisition design, reconstruction algorithms, segmentation, or subsequent processing.
The present study does not yet provide the scale, replication, or independent geometric validation needed for such a benchmark.
Its contribution is to identify the acquisition configurations, reconstruction artifacts, and processing decisions that a field reconstruction dataset should represent and document.  
The results also suggest that unsuccessful or incomplete reconstructions should not automatically be excluded from the dataset.  
Point clouds containing ghost organs, missing surfaces, or registration errors would provide useful examples for developing methods that detect reconstruction failures and estimate uncertainty.  
Preserving both successful and unsuccessful cases would make the collection more representative of field conditions and more informative for evaluating algorithm robustness.
A publicly available dataset built around these principles could help bridge the gap between controlled three-dimensional plant phenotyping and reconstruction from agricultural field imagery.  
It would also provide a common basis for comparing future methods without requiring every research group to repeat the same field acquisitions.  
The potential value of consumer \uav \rgb reconstruction therefore extends beyond producing visually recognizable models of individual plants.  
With documented acquisition conditions, appropriate annotations, and independent validation, these reconstructions could contribute to a reproducible framework for studying plant morphology and developing field phenotyping methods.

\subsection{Study Limitations and Future Validation}

The results should be interpreted within the experimental scope of this study.  
The reconstructions were evaluated through
visual inspection, quantitative point
distribution measurements, and image-based
void analysis, but without independent
three-dimensional ground truth or validated
measurements of individual organs.

The experiments also involved different plants, acquisition trajectories, sampling rates, and field conditions, which were not varied independently in a fully controlled design.  
In particular, the equal frame comparison represents one pair of reconstructions rather than replicated evidence of the general superiority of either acquisition configuration.  
The observed differences therefore identify practical reconstruction behavior under the conditions examined, but do not establish universal relationships between camera geometry, frame count, and geometric accuracy.
The quantitative evaluation of visible reconstruction voids also has limitations.
The void percentages were calculated from selected rendered views rather than directly from the three-dimensional geometry.
Consequently, these measurements depend on the viewing angle, scene boundaries, and the visibility of gaps in the point clouds.
Although they provide a quantitative description of visible reconstruction voids, they do not establish the geometric completeness or accuracy of individual plant organs.
Furthermore, the curlicue experiment was conducted in a different field and growing season, so its void percentage provides a descriptive measure of that reconstruction rather than a controlled comparison with the three earlier acquisition strategies.
Future work should prioritize repeated acquisitions under documented environmental conditions, independent geometric measurements, and quantitative evaluation of organ completeness and reconstruction consistency.  
Controlled comparisons of camera orientation, flight altitude, temporal sampling, and acquisition duration would help identify which factors most strongly affect the reconstruction of different plant organs.  
Automated organ segmentation and uncertainty assessment would then provide a basis for determining whether the resulting geometry supports reliable phenotypic measurements.  
These steps are necessary to move from visually recognizable field reconstructions toward validated three-dimensional plant data suitable for quantitative analysis.

%
%
%
\section{Conclusions}
Consumer \uav \rgb video processed using COLMAP successfully recovered recognizable structures of individual maize organs under field conditions, including leaves, tassels, ears, silks, and brace roots.
Reconstructions of separated plant groups and nursery rows further demonstrated the potential to extend \td reconstruction beyond individual plants, while revealing the increasing challenges of occlusion and preserving fine structural detail.
The comparison using an equal number of frames demonstrated that different image sampling configurations produce distinct point density and isolation characteristics even when the number of frames and source video remain constant.
These findings highlight the importance of considering temporal sampling and viewpoint distribution together when designing UAV acquisitions for \td plant reconstruction.
The masking experiments further showed that preprocessing decisions can affect the preservation of biologically important structures, particularly non-green tissues and fine organs.
Together, these findings provide practical guidance for
acquiring and processing \uav imagery for \td reconstruction
of maize under field conditions.
They also establish a foundation for future work on automated
organ segmentation, geometric validation, and the development
of publicly accessible \td maize datasets for computer vision
and agricultural research.

\section*{Conflict of Interest Statement}
The authors declare that the research was conducted in the absence of any commercial or financial relationships that could be construed as a potential conflict of interest.

\section*{Author Contributions}
Conceptualization, methodology, validation, resources, and writing, D.E.K. and T.K.; software, data curation, and visualization, D.E.K.; supervision, project administration, and funding acquisition, T.K. All authors have read and agreed to the submitted version.

\section*{Funding}
This research received no external funding.

\section*{Acknowledgments}
We thank Chris Browne for field management and Tim Krock, Hadi AliAkbarpour, Kannappan Palaniappan, Dexa Akbar, and Vinny for helpful discussions. We acknowledge support from the Department of Electrical Engineering and Computer Science for D.E.K. and an anonymous gift in aid of maize research.

\section*{Data Availability Statement}
Videos and reconstructions associated with this study are available at \url{https://www.dropbox.com/sh/x4ucxbbjtm6xbmu/AABYrR_V2fFeZQsPhHnB8v5_a?dl=0}.

\bibliographystyle{Frontiers-Harvard}

\end{document}